\documentclass{article}

\usepackage[preprint]{corl_2026} 
\usepackage{graphicx}    
\usepackage{amsmath}     
\usepackage{caption}     
\usepackage{booktabs}
\usepackage{amsmath,amssymb}
\usepackage{float}
\usepackage{caption}
\usepackage[ruled]{algorithm2e}
\usepackage{etoc}
\usepackage{pgffor}
\usepackage{float}
\usepackage{enumitem}
\usepackage{subcaption}
\usepackage{pgffor}
\usepackage{lineno}

\newcommand{\method}{\textsc{Scape}}

\title{\method{}: Scenario-Conditioned Simulation-Augmented Policy Evaluation}

\author{
Dijie Zhu$^{1}$,
Seunghun Oh$^{1,2}$,
Ruopeng Huang$^{1,3}$,
Zhiyu Huang$^{4}$,
Jiaqi Ma$^{1}$,
Chen Tang$^{1}$ \\[0.5em]
$^{1}$University of California, Los Angeles\\
$^{2}$Seoul National University \\
$^{3}$University of Southern California\\
$^{4}$North Carolina State University
}

\begin{document}
\maketitle


\begin{abstract}
Reliable performance evaluation is a central bottleneck for deploying robot-learning policies in real-world conditions. Real-world testing produces faithful evaluation but is costly and difficult to scale, whereas simulation-based testing is easy to scale but inevitably biased due to the sim-to-real gap. Neither source alone is sufficient for scalable and reliable policy evaluation. This motivates simulation-augmented policy evaluation, which combines limited real-world rollouts with abundant simulation proxies for sample-efficient performance estimation. Yet, existing methods focus on estimating performance averaged over initial conditions and deployment settings. By construction, such population-level averages collapse scenario-specific variation, providing limited information about when and where a policy can be safely deployed. To address this limitation, we propose \textbf{\method{}}, a scenario-conditioned simulation-augmented policy evaluation framework that instead predicts scenario-conditioned real-world policy performance from limited paired sim-and-real evaluation samples and large-scale simulation-based testing rollouts. \method{} corrects the sim-to-real bias in the simulation labels before using them to train the prediction model, and calibrates the prediction uncertainty through conformal prediction. We validate \method{} on two embodied task domains, autonomous driving and quadruped velocity tracking. In sim-to-sim settings, it reduces the scenario-level performance prediction error by \textbf{4.9\%}/\textbf{34.7\%} (driving) and \textbf{14.5\%}/\textbf{27.7\%} (quadruped) relative to scene-conditioned neural baselines and aggregate statistical baselines on average. We also validate \method{} for evaluating a velocity-tracking policy deployed on a physical Unitree Go2. In addition, \method{} improves testing sample-efficiency, produces narrower calibrated prediction intervals, enhances generalization to out-of-distribution scenarios, and unlocks fine-grained deployment strategies.
\end{abstract}



\section{Introduction}
\vspace{-5pt}
Recent advances in robot learning, particularly robotic foundation models, have enabled increasingly capable autonomous robots that can operate across a growing range of real-world conditions~\cite{grigorescu2020survey,brohan2022rt,tang2025deep}. A central bottleneck for trustworthy deployment is the rigorous evaluation of learned control policies under diverse operating conditions. Such evaluation typically requires extensive real-world testing on physical hardware, which can be prohibitively expensive, resource-intensive, and safety-critical, especially in open-world environments where robots operate near humans and face rare, and long-tailed conditions. For example, certifying an autonomous vehicle may require billions of miles of driving~\cite{kalra2016driving} under dedicated human oversight. Moreover, relying solely on real-world testing is practically infeasible for covering all rare, long-tailed cases, and the required testing effort keeps growing as the deployment rolls out to new service areas and operating conditions.

\begin{figure}[t]
    \centering
    \includegraphics[width=0.95\linewidth]{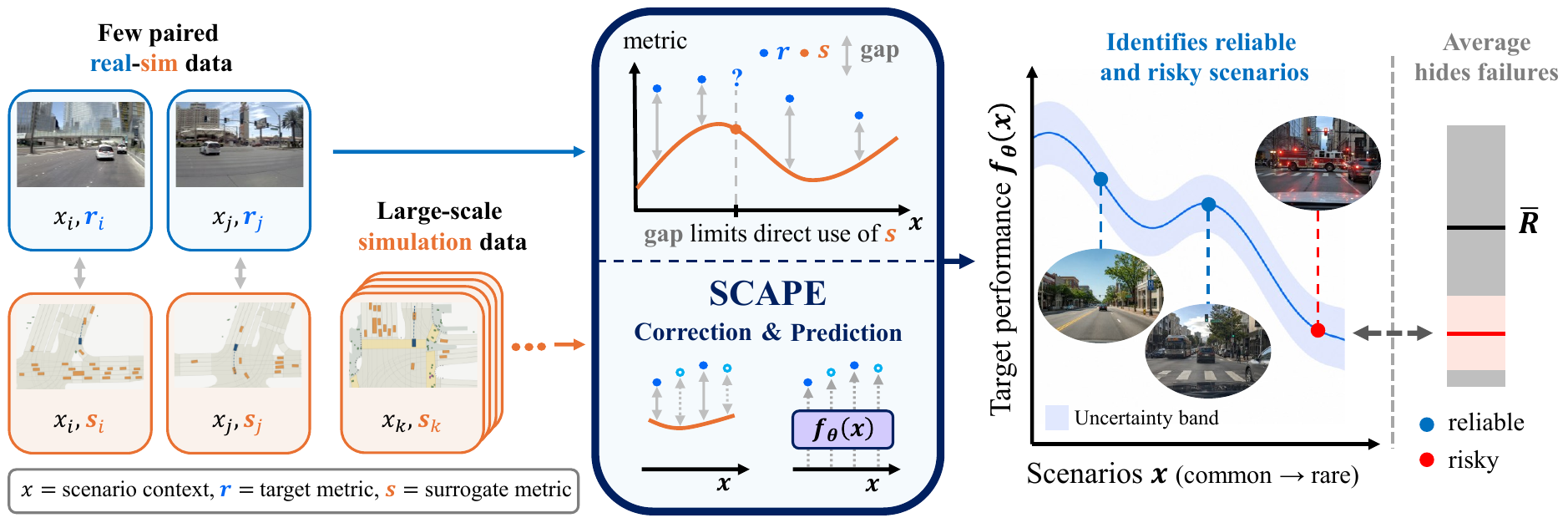}
\caption{ \small
\textbf{Overview of \method{}.}
\method{} combines limited paired sim-and-real data with large-scale simulation data to predict scenario-conditioned target performance with calibrated uncertainty, which supports fine-grained deployment decisions by revealing scenario-specific variation and failures that aggregate metrics can hide.
} 
    \label{fig:headline}
\end{figure}

Simulation offers a scalable alternative, enabling cheap and fast policy rollouts across diverse conditions for scalable evaluation~\citep{mittal2023orbit,li2023scenarionet,gulino2023waymax}. Recent real-to-sim efforts further ground simulators in real-world data~\citep{li2024evaluating,zhang2025real}, improving the correlation between simulation and real-world testing results. Yet, the simulation-to-real gap is inevitable, and the residual bias prevents simulation-only validation from directly certifying real-world deployment. A complementary line of work therefore augments large-scale simulation testing with limited real-world trials to produce calibrated estimates of real-world policy performance, together with valid confidence intervals, through statistical methods such as control variates~\citep{mcbook,luo2025sim2valleveragingcorrelationtest} and prediction-powered inference~\citep{angelopoulos2023prediction,badithela2025suresim}.

However, existing methods primarily estimate the aggregated performance metrics averaged over all deployment conditions. While aggregate estimates can support coarse policy evaluation, much of the information important for deployment decisions is lost through aggregation. A control policy that is strong on average may have varying performance across scenarios and exhibit severe failures in certain cases. Relying solely on the aggregated metrics would withhold deployment until the policy reaches near-perfect average performance, which is impractical. Instead, a sustainable deployment protocol is to gradually roll out deployment into scenarios where the policy is reliable, which facilitates a data flywheel to accumulate real-world rollouts through early deployment to keep improving the policies and gradually expand deployment to broader and more challenging scenarios. Such a protocol requires a fine-grained performance profile of the policy across scenarios and deployment conditions, which existing simulation-augmented policy evaluation methods fail to provide. 

To fill this gap, we instead formulate and study a new \emph{scenario-conditioned simulation-augmented policy evaluation problem}: given limited real-world testing samples and abundant policy rollouts in a simulation, learn a model predicting the real-world policy performance under specific scene context, rather than only the average. As a solution, we propose \textbf{\method{}} ({\textbf{S}cenario-\textbf{C}onditioned Simulation-\textbf{A}ugmented \textbf{P}olicy \textbf{E}valuation), as illustrated in Figure~\ref{fig:headline}. It consists of two modules. First, we train a policy performance prediction model from a combination of limited real-world policy rollouts and large-scale simulation rollouts. Because naively mixing faithful real-world samples with biased simulation samples would sabotage training quality, we adapt the correction-before-augmentation principle from AI-augmented estimation~\citep{wang2026large}: a correction model trained from paired sim-and-real samples first debiases simulation outcomes, the corrected labels then supervise the prediction model training. We further adopt conformal prediction on top of the learned prediction model to provide calibrated uncertainty estimates, allowing developers and decision-makers to identify scenarios where the policy can be confidently deployed or additional real-world testing is necessary.

In summary, our contributions are three-fold: 1) We propose \textbf{\method{}}, a novel framework that combines limited real-world and large-scale simulation testing data for \emph{scenario-conditioned policy evaluation}, addressing the limitations of existing simulation-augmented policy evaluation methods that only estimate the aggregated average performance metrics; 2) We instantiate {\method{}} on two practical applications, autonomous driving and quadruped locomotion, and show through comprehensive experiments that it can effectively leverage simulation to enhance performance prediction accuracy and reduce the required real-world testing samples; 3) We further show that {\method{}} enables various use cases in evaluation and deployment decision-making, such as bootstrapping evaluation on out-of-distribution (OOD) scenarios with simulation and scenario-level policy selection, which are difficult to support when only aggregate performance is estimated.

\section{Related Work}

\paragraph{Simulation and simulation-augmented policy evaluation.} Simulation has become a practical proxy for large-scale policy evaluation when real-world experiments are costly or risky. 
Across embodied simulation benchmarks, platforms such as CARLA~\citep{dosovitskiy2017carla}, MetaDrive~\citep{li2022metadrive}, MuJoCo~\citep{todorov2012mujoco}, and IsaacGym~\citep{makoviychuk2021isaac} provide scalable and reproducible testbeds for evaluating autonomous driving, manipulation, and locomotion policies across diverse conditions. However, simulation-only evaluation remains limited by the sim-to-real gap. Although prior work has attempted to reduce this gap through domain randomization, dynamics randomization, adaptation, and real-to-sim alignment~\citep{tobin2017domain,peng2018sim,tan2018sim,hwangbo2019learning,kumar2021rma,li2024evaluating}, simulation is best viewed as an informative surrogate rather than a direct substitute for real-world validation. This naturally motivates simulation-augmented policy evaluation, which leverages both simulation and limited real-world rollouts to improve evaluation accuracy and sample efficiency. Representative works include Sim2Val~\citep{luo2025sim2valleveragingcorrelationtest}, which exploits correlations across test platforms through control variates, and SureSim~\citep{badithela2025suresim}, which uses limited paired real-simulation evaluations to correct large-scale simulation-based estimates. Although these methods achieve strong estimation performance, they primarily target aggregated mean performance and confidence intervals around the mean. In contrast, \method{} targets scenario-level real-world performance estimation by conditioning on scenario observations, enabling fine-grained estimates.



\vspace{-10pt}
\paragraph{Uncertainty quantification for policy evaluation.}
Policy comparison and selection require more than point estimates, since apparent performance differences can be dominated by evaluation noise when trials are limited. Single point estimates also do not quantify the variability of policy outcomes across trials or scenarios, limiting their usefulness for downstream decision-making. Prior work has constructed confidence intervals directly from real-world evaluations, including statistical procedures for policy success rates \citep{vincent2024generalizable} and classical concentration inequalities for bounded performance metrics \citep{hoeffding1963probability}. However, real-data-only intervals often require many costly trials to become informative, making them impractical. To reduce the real-world data burden, later work often leverages statistical frameworks, as in SureSim \citep{badithela2025suresim} and Sim2Val \citep{luo2025sim2valleveragingcorrelationtest}, to provide confidence intervals or variance-reduced estimates. However, these intervals are primarily designed for aggregated mean estimates, they quantify uncertainty in the population mean rather than the conditional variability of an incoming scenario. In contrast, \method{} incorporates conformal prediction~\citep{lei2018distribution,angelopoulos2023gentle} to produce calibrated uncertainty intervals for scenario-level estimates.

\begin{figure}[t]
    \centering
    \includegraphics[
        width=0.95\linewidth,
        height=0.5\textheight,
        keepaspectratio
    ]{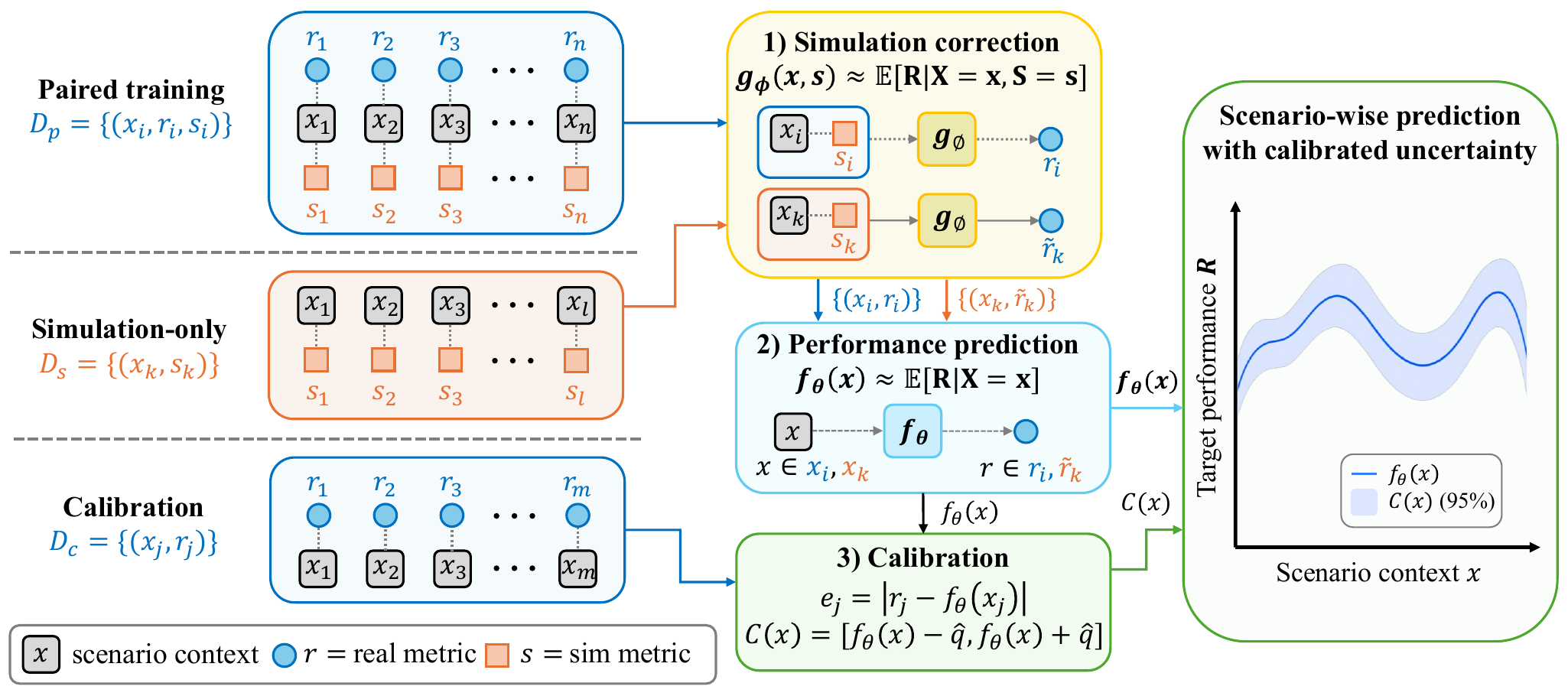}
    \caption{ \small
\textbf{Method overview of \method{}.}
\method{} corrects biased simulation labels using paired sim-and-real data, trains a scenario-conditioned performance predictor with corrected simulation-only data, and applies conformal calibration to quantify prediction uncertainty.
}\label{fig:scape_overview}
\end{figure}


\section{Scenario-Conditioned Simulation-Augmented Policy Evaluation}
\label{sec:method}
\vspace{-6pt}
This section formalizes the \emph{scenario-conditioned simulation-augmented policy evaluation} problem (Sec.~\ref{subsec:problem}) and presents \method{}, our solution framework  for correcting biased simulation labels, training a scenario-level predictor, and calibrating uncertainty with conformal prediction (Sec.~\ref{subsec:framework}). 

\subsection{Problem Formulation}
\label{subsec:problem}
We study how to predict the real-world performance of a given policy in a \emph{specific} scenario using limited real-world testing and abundant simulation-based testing. Existing simulation-augmented policy evaluation methods such as Sim2Val~\citep{luo2025sim2valleveragingcorrelationtest} and SureSim~\citep{badithela2025suresim} estimate an \emph{aggregate} quantity, namely the policy's mean performance over the entire scenario distribution, together with a confidence interval around the estimated mean. An aggregate estimate supports coarse evaluation and ranking, but hides scenario-specific variation and failures. We therefore target the scenario-conditioned real-world performance and provide a calibrated uncertainty interval for each scenario-level prediction.

Formally, let \(X\in\mathcal{X}\) denote a \emph{scenario}, represented by the scene context and initial conditions under which the policy is evaluated. For a scenario $X$, let $R$ be the policy's real-world performance under an evaluation metric of interest, either continuous or binary, and $S$ the corresponding metric measured through simulation testing. Both measure the same evaluation metric on the same scenario, with \(R\) trusted but expensive to estimate and \(S\) cheap but biased. Evaluating the policy in both worlds on the same scenario yields a \emph{paired} sample $(X,R,S)$, whereas evaluating it in simulation alone yields a \emph{simulation-only} sample $(X,S)$. We assume access to a small set of paired samples with abundant simulation-only samples, which are aggregated into three disjoint datasets: 
\[
\underbrace{\mathcal{D}_{p}=\{(x_i,r_i,s_i)\}_{i=1}^{n}}_{\text{paired training set}},
\qquad
\underbrace{\mathcal{D}_{s}=\{(x_k,s_k)\}_{k=1}^{\ell}}_{\text{simulation-only training set}},
\qquad
\underbrace{\mathcal{D}_{c}=\{(x_j,r_j)\}_{j=1}^{m}}_{\text{held-out calibration set}},\quad \text{with}\ \  {\ell\gg m+n}.
\]

Our objective is two-fold. First, we train a model predicting the scenario-conditioned real-world policy performance from the paired and simulation-only training data $\mathcal{D}_p\cup\mathcal{D}_s$,
$
f_\theta(x)\;\approx\;\mu(x):=\mathbb{E}[\,R\mid X=x\,],
$
which is the conditional mean for a continuous metric and the positive-class probability $\mathbb{P}(R=1\mid X=x)$ for a binary one. Second, given the held-out calibration set $\mathcal{D}_c$ and a miscoverage level $\alpha\in(0,1)$, we equip each prediction with an uncertainty region $C_\alpha(x)$\textemdash an interval for continuous metrics and a set for binary ones\textemdash such that $\mathbb{P}\!\left(R\in C_\alpha(X)\right)\;\ge\;1-\alpha,$
where the probability is over the exchangeable draw of the calibration and test scenarios. A wider $C_\alpha(x)$ reflects a less certain prediction and flags a scenario requiring further real-world testing.




\subsection{\method{} Framework}
\label{subsec:framework}
A central challenge in simulation-augmented policy evaluation is exploiting large-scale simulation data without inheriting its sim-to-real bias. Since simulation rollouts are cheap but biased, naively mixing them with real-world data can pull predictions toward the simulation distribution. Prior work addresses this challenge through \emph{prediction-powered inference} (PPI)~\cite{angelopoulos2023prediction,badithela2025suresim}, using paired real-simulation evaluations to debias aggregate metric estimates.
In principle, PPI could be adapted to train our scenario-conditioned predictor \(f_\theta\). In practice, \(f_\theta\) must often be a neural network because robotic scenario contexts are high-dimensional, such as HD-map and agent-history features in autonomous driving or elevation maps in quadruped locomotion (Sec.~\ref{sec:experiment-setting}). Yet, in our experiments, adapting PPI to high-dimensional, non-convex neural training, where its guarantees no longer directly apply, is unstable and provides little benefit from paired data (Sec.~\ref{sec:result}). We therefore propose {\bf \method{}}, which adopts an alternative bias-correction strategy that works reliably well for training the neural scenario-conditioned policy performance prediction model $f_\theta$. The detailed algorithm of \method{} is summarized in the Appendix (see Algorithm~\ref{alg:scape}).

{\bf Simulation-augmented policy performance prediction.} We instead build {\method{}} on an alternative bias-correction principle adopted from AI-augmented estimation (AAE) \citep{wang2026large}.  Originally developed for combining scarce human labels with abundant LLM-generated surrogate labels for human-choice modeling, AAE first corrects surrogate labels using paired data and then augments model training with the corrected data. We adapt this strategy to train neural, scenario-conditioned performance prediction models and find it works remarkably well in our experiments.

Concretely, \method{} trains the performance prediction model $f_\theta$ in two stages (Figure~\ref{fig:scape_overview}). First, we train a \emph{calibration model} $g_\phi$ on the paired set $\mathcal{D}_p$ to predict the real-world metric from a scenario together with its simulation result,
$
g_\phi(x,s)\;\approx\;\mathbb{E}[\,R\mid X=x,\,S=s\,],
$ 
where conditioning on $x$ lets $g_\phi$ correct the sim-to-real bias based on scenario rather than applying a single global offset. We then apply $g_\phi$ to every simulation-only sample $(x_k,s_k)\in\mathcal{D}_s$ to obtain a corrected label $\tilde r_k=g_\phi(x_k,s_k)$, turning the abundant simulation-only data into a large corrected set $\mathcal{D}_{\mathrm{pseudo}}=\{(x_k,\tilde r_k)\}_{k=1}^{\ell}$. In the second stage, we train the scenario-conditioned prediction model $f_\theta$ on the union of the trusted real-world labels and the corrected simulation-only labels,
\[
\mathcal{D}_{\mathrm{aug}}=\{(x_i,r_i)\}_{i=1}^{n}\;\cup\;\mathcal{D}_{\mathrm{pseudo}},
\qquad
\hat\theta\;\in\;\arg\min_{\theta}\;\frac{1}{n+\ell}\!\!\sum_{(x,r)\in\mathcal{D}_{\mathrm{aug}}}\!\!\mathcal{L}\big(f_\theta(x),\,r\big),
\]
where $\mathcal{L}$ is the squared error for continuous metrics and the cross-entropy for discrete metrics. The final prediction model $f_\theta$ takes only the scenario context $x$ as input without the surrogate metric $s$. This design enables predicting real-world performance at deployment without requiring access to a simulator for collecting additional simulation rollouts given a scenario of interest. 


{\bf Uncertainty quantification via conformal prediction.} The prediction model $f_\theta$ provides scenario-conditioned point prediction, but trustworthy deployment decisions also require quantifying the uncertainty of the prediction. We incorporate conformal prediction~\citep{lei2018distribution,angelopoulos2023gentle} into \method{} to construct the calibrated uncertainty set $C_\alpha(x)$. Conformal prediction is model-free and, under exchangeability of the calibration and test scenarios, provides finite-sample marginal coverage $\mathbb{P}(R\in C_\alpha(X))\ge 1-\alpha$ with no distributional assumptions on the prediction model $f_\theta$ or the data. On the held-out calibration set $\mathcal{D}_c=\{(x_j,r_j)\}_{j=1}^{m}$, we score each scenario against its real-world label $r_j$. For a continuous metric, we compute the calibration residual $e_j=|r_j-f_\theta(x_j)|$ for each \((x_j,r_j)\in\mathcal{D}_c\), sort the scores $e_{(1)}\le\cdots\le e_{(m)}$, set $\hat q_{1-\alpha}=e_{(\lceil(m+1)(1-\alpha)\rceil)}$, and return the interval $C_\alpha(x)=[\,f_\theta(x)-\hat q_{1-\alpha},\;f_\theta(x)+\hat q_{1-\alpha}\,]$. For a discrete metric we compute the analogous classification conformity scores on $\mathcal{D}_c$ and return a calibrated prediction set. Note that this split conformal prediction constructs the interval $C_\alpha(x)$ with a scenario-invariant half-width $\hat q_{1-\alpha}$. The resulting intervals therefore do not provide scenario-dependent widths or conditional coverage. Nevertheless, the interval center $f_\theta(x)$ still captures variations across scenarios, which allow the calibrated uncertainty sets to support fine-grained deployment decisions, for example, relative to a performance threshold, an interval can certify deployment, reject deployment, or flag scenarios for additional testing.

\vspace{-4pt}
\section{Experiments}
\begin{figure}[t]
    \centering
    \includegraphics[
        width=0.98\linewidth,
        height=0.4\textheight,
        keepaspectratio
    ]{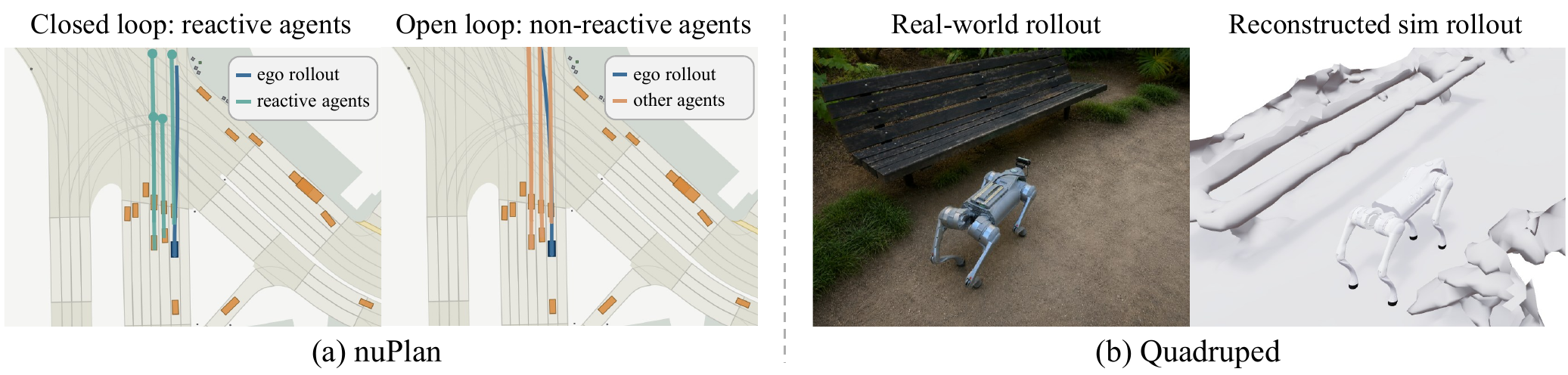}
    \vspace{-2mm}
    \caption{\small \textbf{Paired sample construction in (a) nuPlan and (b) quadruped experiments.}
In nuPlan, each pair is constructed from the same driving scenario by measuring the target metric $R$ with closed-loop evaluation and the surrogate metric $S$ with open-loop evaluation.
In quadruped, each pair matches a real-world rollout with a reconstructed-sim rollout under the same scene context, robot placement, and command.}
    \label{fig:Paired-env}
    \vspace{-4mm}
\end{figure}


\subsection{Experimental Settings}
\label{sec:experiment-setting}
We evaluate \method{} in two embodied domains---autonomous driving and quadruped velocity tracking---shown in Figure~\ref{fig:Paired-env}. We first study both domains in a \emph{Sim2Sim} setting, treating outcomes from a higher-fidelity or perturbed simulator as target labels \(R\) and those from a cheaper or nominal simulator as surrogate measurements \(S\). In each domain, we construct a paired training set \(\mathcal{D}_p\) and a simulation-only training set \(\mathcal{D}_s\) for training the point-prediction models. In the experiments, we vary the fractions of \(\mathcal{D}_p\) and \(\mathcal{D}_s\) used for training to study how prediction accuracy scales with paired and surrogate data. We additionally reserve three paired splits: a validation set \(\mathcal{D}_v\) for neural-model selection, a calibration set \(\mathcal{D}_c\) for split-conformal calibration, and a test set \(\mathcal{D}_{\mathrm{test}}\) for final evaluation. We further evaluate quadruped velocity tracking in \emph{Sim2Real} using real-world rollouts collected from a physical Unitree Go2 across multiple indoor and outdoor scenarios.


{\bf Autonomous driving (nuPlan).} We conduct experiments on nuPlan~\citep{caesar2021nuplan} and evaluate three ML-based planners, \texttt{urban\_driver\_open\_loop\_model}, \texttt{vector\_model}, and \texttt{simple\_vector\_model} provided by nuPlan, each trained under their default configurations. The scenario feature \(x\) is extracted by a frozen \texttt{UrbanDriver} encoder. We record four evaluation metrics: average displacement error (ADE), time-to-collision under 1\,s (TTC\,\(<1\)s), drivable-area compliance, and no ego at-fault collision (see Appendix for their definition). The surrogate metric \(S\) comes from fast open-loop evaluation with non-reactive agents, while the target metric \(R\) comes from closed-loop evaluation with reactive agents. Detailed dataset statistics and feature composition are provided in Appendix~\ref{app:nuplan_dataset_stats}.


{\bf Quadruped velocity tracking.} We evaluate an RL-trained quadruped velocity-tracking policy in both \emph{Sim2Sim} and \emph{Sim2Real} settings. Each scenario feature \(x\) consists of local terrain geometry, a linear velocity command, and a yaw-rate command. Each 2-s rollout produces two target metrics: velocity-tracking error \(\mathrm{MAE}_{\mathrm{vel}}\) and yaw-rate-tracking error \(\mathrm{MAE}_{\mathrm{yaw}}\). In Sim2Sim, \(S\) is measured in a nominal simulator and \(R\) in a perturbed simulator with dynamics randomization and external disturbances. In Sim2Real, \(R\) is collected from physical Go2 rollouts and \(S\) from matched simulation rollouts. We construct paired real-sim scenes using the GaussGym~\citep{escontrela2025gaussgym} real-to-sim pipeline and align robot poses with HLoc~\citep{sarlin2019coarse} and SuperGlue~\citep{sarlin2020superglue}. Dataset details are provided in Appendix~\ref{app:locomotion_dataset_stats}.

{\bf Baselines.} \method{} addresses a new setting without direct off-the-shelf baselines, so we compare it against two families of adapted methods. The first family consists of \emph{aggregate-mean estimators}, adapted from prior simulation-augmented policy evaluation methods~\cite{luo2025sim2valleveragingcorrelationtest,badithela2025suresim}, that return a single scenario-independent mean estimate: 1)~\textbf{MC} (Monte Carlo) estimates the mean from the real-world labels only~\citep{mcbook}; 2)~\textbf{CV} (Control Variates) treats the simulation metric as a control variate~\citep{mcbook,luo2025sim2valleveragingcorrelationtest}; 3)~\textbf{PPI} (Prediction-Powered Inference) debiases surrogate-based mean estimates using paired samples~\citep{angelopoulos2023prediction,badithela2025suresim}. The second family consists of \emph{scenario-conditioned neural predictors} \(f_\theta(x)\): 4)~\textbf{R-Only} trains on the real-world labels only; 5)~\textbf{RS-Mix} naively mixes real-world and simulation labels; 6)~\textbf{PPI-NN} trains with the rectified PPI objective~\citep[Sec.~4.3]{angelopoulos2023prediction}. For a fair comparison, all baselines and \method{} use the same paired and surrogate data budgets to train the point-prediction models. After training, we apply the same split-conformal procedure, using the same calibration set \(\mathcal{D}_c\), to all methods to isolate the effect of the point-prediction model on the resulting prediction intervals.\footnote{Native PPI and CV confidence intervals target the population mean rather than the outcome for a new scenario, so we use only the mean-estimation components of these baselines in our experiments.} The hyperparameters and optimization protocol for the neural baselines and \method{} are detailed in Appendix~\ref{tab:mlp-hyperparameters}.

\subsection{Experimental Results}
\label{sec:result}

\begin{figure}[t]
    \centering
    \vspace{-6pt}
    \includegraphics[
      width=0.95\linewidth,
      height=0.33\textheight,
      keepaspectratio
    ]{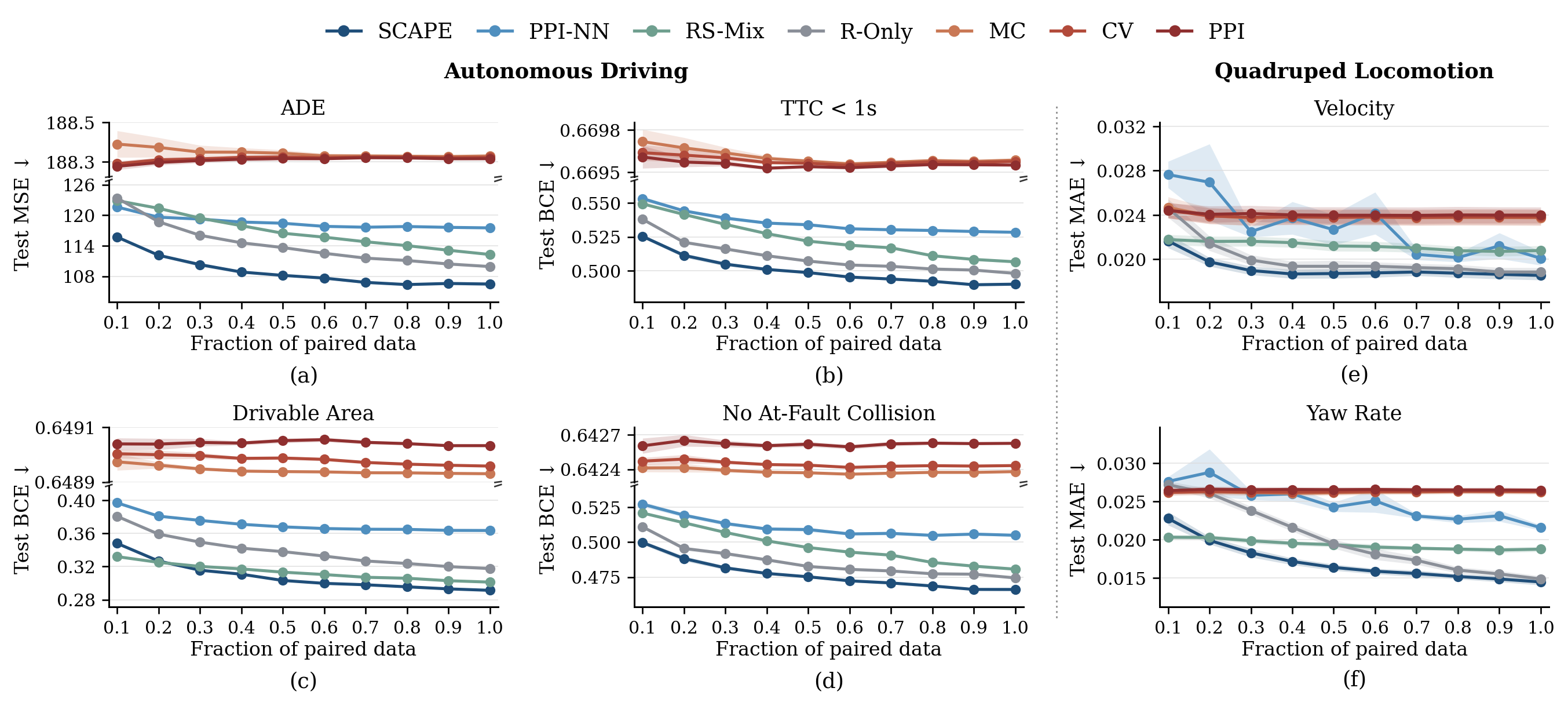}
    \caption{\small \textbf{Test loss vs. fraction of paired-label on autonomous driving and quadruped locomotion tasks.}
    The paired-data fraction denotes the fraction of the full paired training set \(\mathcal{D}_p\) used for training.
    The Urban Driver setting contains \(83.6\)K paired training samples and \(119.2\)K surrogate-only samples; the quadruped Sim2Sim setting contains \(300\) paired training samples and \(1{,}493\) surrogate-only samples. Panels (a)--(d) report autonomous driving Sim2Sim results, and panels (e)--(f) report quadruped Sim2Sim velocity-tracking results. Curves report mean \(\pm\) standard error over 10 random paired-data seeds.}
    \label{fig:main_results}
    \vspace{-5pt}
\end{figure}

Figure~\ref{fig:main_results} reports Sim2Sim scenario-level prediction errors as the paired-data fraction varies for nuPlan and quadruped velocity tracking\footnote{Figure~\ref{fig:main_results} (a)-(d) shows the results for \texttt{UrbanDriver}. The results for the other planners are in Appendix~\ref{app:nuplan_main_results}.}. In both domains, scenario-conditioned neural methods outperform aggregate-mean estimators, confirming the need for fine-grained evaluation. More importantly, \method{} is the only method that consistently outperforms R-Only across paired-data fractions and metrics. In contrast, RS-Mix and PPI-NN often fail to exploit low-cost simulation effectively, as biased surrogate labels can degrade prediction accuracy. Across the complete evaluation suite reported in the main paper and appendix, \method{} reduces prediction errors by $4.9\%/34.7\%$ on nuPlan and $14.5\%/27.7\%$ on quadruped tracking relative to the scenario-conditioned neural/aggregate statistical baselines.  These averages aggregate all evaluated planners, settings, paired-data fractions, and target metrics. Compared with the strongest baseline, the corresponding gains are $2.0\%$ and $4.08\%$. It also matches the strongest baseline full-budget performance with \(20\%\)--\(60\%\) fewer paired labels on nuPlan and \(10\%\)--\(60\%\) fewer on quadruped velocity tracking. More importantly, \method{} is the \emph{only surrogate-augmented method that consistently improves on R-Only}. It shows that simply enlarging the training set with biased surrogate labels does not reliably lead to benefits in prediction accuracy and sample efficiency. The consistent gains of \method{} therefore reflect its ability to correct those labels and turn them into useful supervision. These gains are also practically meaningful: even increasing the amount of real labeled data available to R-Only by \(10\times\) reduces its prediction error by only \(8.81\%\) on average.

For uncertainty quantification, Table~\ref{tab:conformal-main-combined} reports conformal widths at paired-label fraction \(1.0\). \method{} achieves the narrowest width in all six metric settings,\footnote{Full paired-data sweeps and empirical coverage results are in Appendix~\ref{app:conformal full}.} while aggregate-mean estimators often produce large or full prediction sets. Thus, \method{} improves prediction accuracy while providing more efficient calibrated uncertainty at valid coverage.

\begin{table}[H]
\centering
\vspace{-6pt}
\captionsetup{skip=8pt}
\caption{\small \textbf{Split conformal width at nominal coverage $1-\alpha=0.95$.} Results for fraction of paired-label $1.0$ are reported. For the nuPlan regression target ADE, width is the prediction-interval length $2\hat q$ in meters; for the three nuPlan binary targets, width is the average conformal prediction-set size $|C(x)|\in[0,2]$. For quadruped Sim2Sim, width is the regression interval length reported as $\times 10^{-2}$. Cells are mean~$\pm$~standard error over 10 paired-data seeds. \textbf{Bold} marks the narrowest width per column.}
\label{tab:conformal-main-combined}
\small
\setlength{\tabcolsep}{4.6pt}
\renewcommand{\arraystretch}{1.18}
\resizebox{0.8\textwidth}{!}{%
\begin{tabular}{l|c|c|c|c|c|c}
\toprule
& \multicolumn{4}{c|}{\textbf{nuPlan Urban Driver}} & \multicolumn{2}{c}{\textbf{Go2 Sim2Sim}} \\
\cmidrule(lr){2-5}\cmidrule(lr){6-7}
Method & ADE & TTC$<1$s & Drivable & Collision & Velocity & Yaw Rate \\
\midrule
\method{} & $\boldsymbol{40.07\!\pm\!0.26}$ & $\boldsymbol{1.447\!\pm\!0.010}$ & $\boldsymbol{1.118\!\pm\!0.004}$ & $\boldsymbol{1.430\!\pm\!0.004}$ & $\boldsymbol{10.84\!\pm\!0.48}$ & $\boldsymbol{7.60\!\pm\!0.39}$ \\
ppi-nn & $44.44\!\pm\!0.16$ & $1.614\!\pm\!0.004$ & $1.233\!\pm\!0.004$ & $1.573\!\pm\!0.004$ & $11.42\!\pm\!0.38$ & $10.27\!\pm\!0.31$ \\
rs-mix & $44.65\!\pm\!0.16$ & $1.505\!\pm\!0.006$ & $1.152\!\pm\!0.002$ & $1.495\!\pm\!0.003$ & $11.56\!\pm\!0.28$ & $9.12\!\pm\!0.22$ \\
r-only & $41.02\!\pm\!0.30$ & $1.497\!\pm\!0.003$ & $1.155\!\pm\!0.003$ & $1.474\!\pm\!0.009$ & $11.12\!\pm\!0.59$ & $7.75\!\pm\!0.31$ \\
mc & $50.27\!\pm\!0.00$ & $2.000\!\pm\!0.000$ & $2.000\!\pm\!0.000$ & $2.000\!\pm\!0.000$ & $12.64\!\pm\!0.88$ & $11.95\!\pm\!0.31$ \\
cv & $50.29\!\pm\!0.00$ & $2.000\!\pm\!0.000$ & $2.000\!\pm\!0.000$ & $2.000\!\pm\!0.000$ & $12.50\!\pm\!0.88$ & $11.88\!\pm\!0.29$ \\
ppi & $50.31\!\pm\!0.00$ & $2.000\!\pm\!0.000$ & $2.000\!\pm\!0.000$ & $2.000\!\pm\!0.000$ & $12.41\!\pm\!0.84$ & $11.75\!\pm\!0.29$ \\
\bottomrule
\end{tabular}%
}
\vspace{-5pt}
\end{table}

\paragraph{Sim2Real quadruped velocity tracking.} We conduct the quadruped velocity tracking using policy rollouts collected on a physical Unitree Go2 robot. We collected data by placing the robot at diverse positions across five different scenes, including flat and sloped terrains\footnote{Detailed scene structures and reconstructed pairs are provided in Appendix~\ref{app:locomotion_dataset_stats}.}. As shown in Table~\ref{tab:real2sim-go2}, \method{} achieves the lowest MAE on both target metrics, reducing error by \(11.3\%\) on \(\mathrm{MAE}_{\mathrm{vel}}\) and \(8.7\%\) on \(\mathrm{MAE}_{\mathrm{yaw}}\) compared with the strongest baseline. These results indicate that \method{} remains effective beyond Sim2Sim evaluation and can effectively facilitate robot policy evaluation and deployment in the physical environments. 
\begin{table}[!t]
\centering
\vspace{-10pt}
\captionsetup{skip=8pt}
\caption{\textbf{Quadruped Sim2Real velocity tracking.}
Test MAE on the sim-to-real velocity tracking task. Cells are mean $\pm$ standard error over 10 random paired-data seeds.
\textbf{Bold} marks the lowest error per row.}
\label{tab:real2sim-go2}
\scriptsize
\setlength{\tabcolsep}{6pt}
\renewcommand{\arraystretch}{1.15}
\resizebox{\textwidth}{!}{%
\begin{tabular}{l|ccccccc}
\toprule
\textbf{Metric ($\times 10^{-3}$)} & mc & cv & ppi & r\_only & rs\_mix & ppi-nn & \method{} \\
\midrule
$R_{\mathrm{vel}}$ test MAE  & $16.55\!\pm\!0.96$ & $17.07\!\pm\!0.86$ & $26.51\!\pm\!1.52$ & $13.29\!\pm\!0.76$ & $16.92\!\pm\!0.99$ & $19.16\!\pm\!1.79$ & $\boldsymbol{11.79\!\pm\!0.70}$ \\
$R_{\mathrm{yaw}}$ test MAE & $5.24\!\pm\!0.41$ & $5.48\!\pm\!0.41$ & $6.86\!\pm\!0.56$ & $5.16\!\pm\!0.41$ & $6.22\!\pm\!0.47$ & $5.37\!\pm\!0.45$ & $\boldsymbol{4.71\!\pm\!0.32}$ \\
\bottomrule
\end{tabular}
}
\vspace{-10pt}
\end{table}

\subsection{Additional Experiments and Analyses on nuPlan}
We conduct additional nuPlan experiments to analyze and highlight several key characteristics of \method{}. Figures~\ref{fig:urban-ood-correction} and~\ref{fig:decision-making} show the results on \texttt{UrbanDriver} with full paired data, and the complete results are provided in Appendix~\ref{app:ood results} and ~\ref{app:correction results}.
\paragraph{\method{} scales with surrogate data where baselines saturate.}
We examine how prediction accuracy changes as surrogate-only data increases. We sweep the surrogate data fractions given fixed paired-data budget for the three scenario-conditioned methods. As shown in Figure~\ref{fig:urban-ood-correction}, \method{} keeps improving as more cheap surrogate data is incorporated. In contrast, RS-Mix peaks early and then degrades because naive mixing pollutes the training signal. PPI-NN also benefits from additional surrogate data but exhibits higher empirical training variance and instability. We hypothesize that this behavior arises because its subtractive correction is estimated from limited paired data, which can induce conflicting, high-variance gradients during non-convex neural-network training. In contrast, \method{} separates correction from augmentation. It first converts surrogate data into target-aligned pseudo-labels and then trains the final predictor separately.

\paragraph{\method{} benefits performance prediction on out-of-distribution scenarios.} Deployment often expands from familiar scenarios to broader OOD conditions, requiring performance estimates before target testing. We test whether \method{} can use surrogate-only data from reconstructed OOD scenarios to improve prediction without OOD target labels. In nuPlan, we leave out one city from paired training and evaluate on the test set (Figure~\ref{fig:urban-ood-correction}). Across all evaluated planners, without OOD surrogate data, \method{} already achieves the lowest average loss, outperforming the other scenario-conditioned baselines by \(3.3\%\); adding surrogate-only data from the held-out city further reduces its loss by \(4.8\%\), recovering \(95.3\%\) of full-training performance and widening its margin over the scenario-conditioned baselines to \(5.3\%\). Note that this experiment evaluates OOD point-prediction performance; we do not claim conformal coverage under this distribution shift.

\begin{figure}[!b]
    \centering
    \captionsetup{skip=3pt}
    \includegraphics[
      width=\linewidth,
      height=0.3\linewidth,
    ]{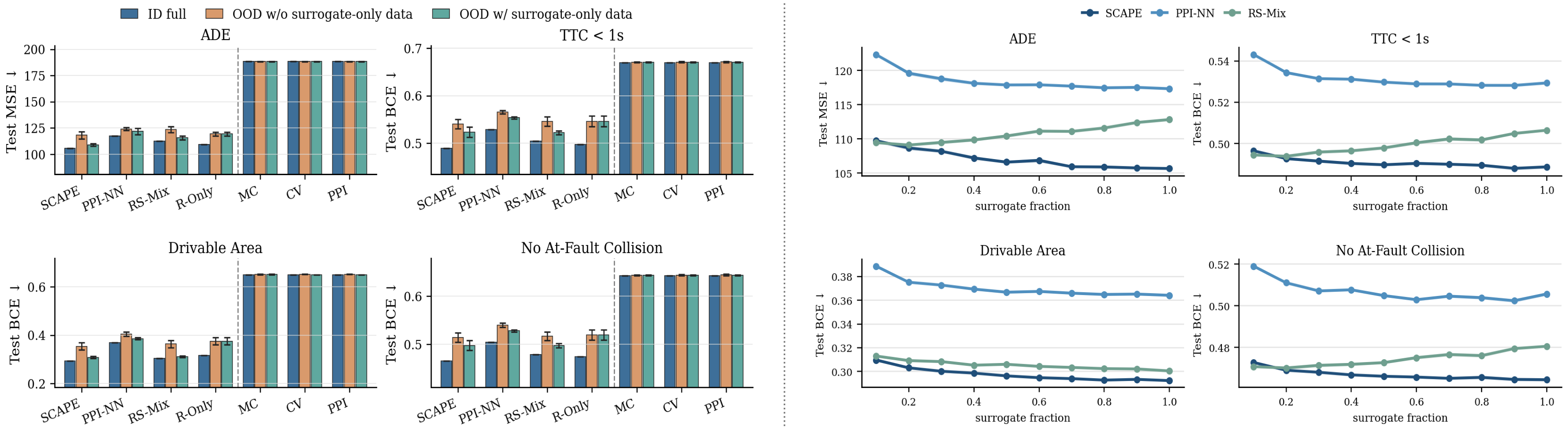}
    \caption{\small \textbf{Additional nuPlan results.} Left: city-level OOD generalization results for \texttt{UrbanDriver}. Bars show mean \(\pm\) standard error over 4 city holdouts. Right: \texttt{UrbanDriver} results at paired data fraction \(1.0\) with varying surrogate-only fraction. Curves report mean \(\pm\) standard error over 9 (3 paired-data seed \(\times\) 3 surrogate-data seed) random seeds.}

    \label{fig:urban-ood-correction}
    \vspace{-15pt}
\end{figure}

\begin{figure}[H]
    \centering
    \includegraphics[
      width=0.89\linewidth,
      keepaspectratio
    ]{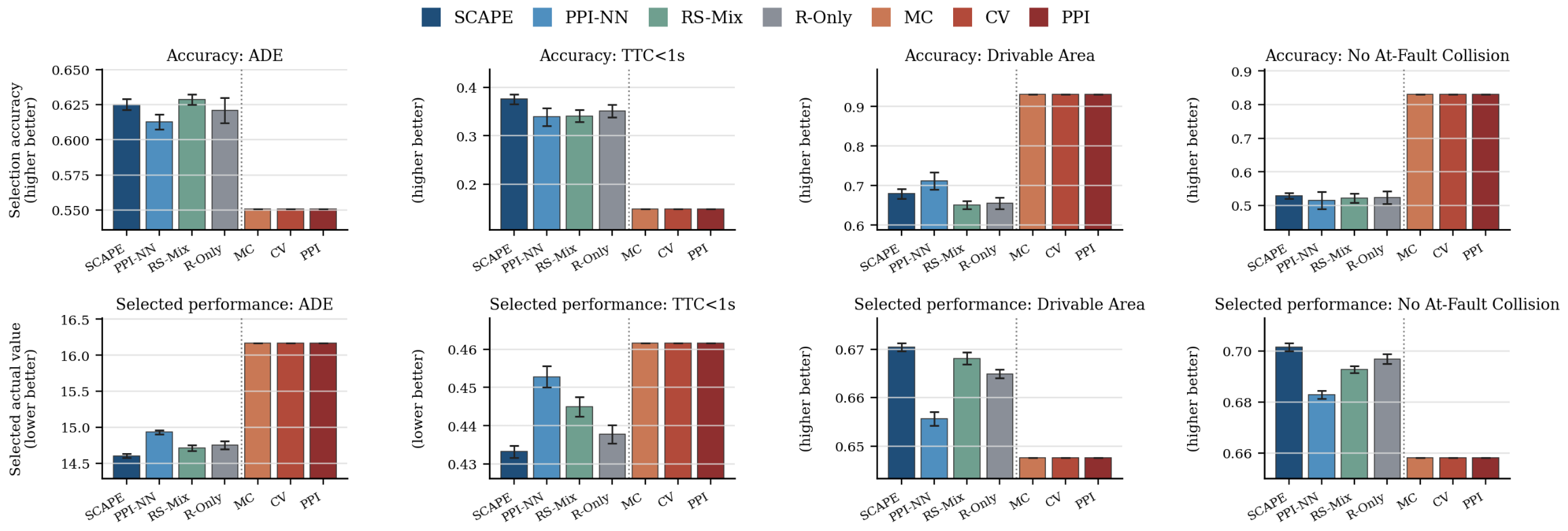}
    \caption{\small \textbf{Evaluator guided planner selection accuracy and corresponding selected performance.}
    Results for fraction of paired-label $1.0$ are reported. Learned evaluators are used to select the predicted best planner for each scenario. Bars show mean \(\pm\) standard error over 10 paired-data seeds.}
    \label{fig:decision-making}
    \vspace{-10pt}
\end{figure}

\vspace{-6pt}
\paragraph{Scenario-level evaluation enables scenario-conditioned deployment strategy.} \method{} enables scenario-conditioned policy selection instead of using a single policy universally. To demonstrate this benefit, we use each evaluator as a router that selects the predicted best planner for each scenario. Figure~\ref{fig:decision-making} shows that \method{} achieves the best selected performance on all four metrics, improving over the scenario-conditioned and aggregate-mean baselines by \(1.67\%\) and \(6.47\%\) on average, respectively. Although aggregate-mean baselines show high selection accuracy on drivable-area compliance and no ego at-fault collision, they often route all scenarios to the best-average planner, masking rare failures. In contrast, \method{} often selects a near-best planner even when it misses the oracle best, yielding better ensemble performance.





\vspace{-6pt}
\section{Conclusion}
\vspace{-6pt}
\label{sec:conclusion}
In this work, we present \method{}, a scenario-conditioned, simulation-augmented framework for policy evaluation that estimates scenario-conditioned real-world policy performance with calibrated uncertainty from limited paired sim-and-real evaluations and large-scale simulation-only rollouts. Across autonomous driving and quadruped velocity tracking, and validated from simulation to a physical Unitree Go2, \method{} predicts real-world performance more accurately than scene-conditioned neural and aggregate statistical baselines, uses scarce real-world testing samples more efficiently, generalizes better to OOD scenarios, and produces narrower calibrated intervals. It also enables fine-grained, scenario-level policy comparison and deployment strategy.

{\bf Limitations and future work.}
\method{} currently uses split conformal prediction, which provides an uncertainty set with identical size over the scenario distribution, instead of true conditional coverage with scenario-dependent width. Our real-world evaluation also remains limited in scope, and broader experiments across different embodiments, tasks, and a broader range of scenarios are helpful to fully characterize \method{}'s performance. Finally, while \method{} yields strong empirical performance, rigorous theoretical analysis would further strengthen its value for trustworthy policy deployment.


\clearpage



\bibliography{reference}  

\clearpage
\appendix
\raggedbottom
\section{Appendix}
\label{app:main}

\etocsettocstyle{\subsection*{Appendix Contents}}{}
\localtableofcontents
\clearpage
\subsection{\method{} Algorithm Details}
\label{app:algorithm}
\begin{nolinenumbers}
\begin{algorithm}[H]
\small
\DontPrintSemicolon
\SetAlgoNoLine
\caption{\method{}}
\label{alg:scape}

\KwData{
Paired training set 
\(\mathcal{D}_{p}=\{(x_i,r_i,s_i)\}_{i=1}^{n}\), 
surrogate-only set 
\(\mathcal{D}_{s}=\{(x_k,s_k)\}_{k=1}^{\ell}\), 
calibration set 
\(\mathcal{D}_{c}=\{(x_j,r_j)\}_{j=1}^{m}\), 
miscoverage level \(\alpha\)
}

\KwResult{
Scenario-level estimator \(f_\theta\) and calibrated prediction interval \(C_\alpha(x)\)
}

\vspace{1mm}

\newcounter{scapealgline}
\setcounter{scapealgline}{0}
\newcommand{\algnum}{%
    \refstepcounter{scapealgline}%
    \makebox[1.8em][r]{\textbf{\arabic{scapealgline}:}}\hspace{0.45em}%
}

\noindent
\begin{minipage}[t]{0.60\linewidth}
\algnum \(g_\phi \leftarrow \textsc{TrainCorrection}(\mathcal{D}_p)\)\;

\algnum \textbf{for} \((x_k,s_k)\in \mathcal{D}_s\) \textbf{do}\;

\algnum \hspace*{1.5em}\(\tilde r_k \leftarrow g_\phi(x_k,s_k)\)\;

\algnum \textbf{end}\;

\algnum \(\mathcal{D}_{\mathrm{pseudo}} \leftarrow \{(x_k,\tilde r_k)\}_{k=1}^{\ell}\)\;

\algnum \(\mathcal{D}_{p}^{R} \leftarrow \{(x_i,r_i)\}_{i=1}^{n}\)\;

\algnum \(\mathcal{D}_{\mathrm{aug}} \leftarrow \mathcal{D}_{p}^{R}\cup\mathcal{D}_{\mathrm{pseudo}}\)\;

\algnum \(f_\theta \leftarrow \textsc{TrainEvaluator}(\mathcal{D}_{\mathrm{aug}})\)\;

\algnum \textbf{for} \((x_j,r_j)\in \mathcal{D}_c\) \textbf{do}\;

\algnum \hspace*{1.5em}\(e_j \leftarrow |r_j-f_\theta(x_j)|\)\;

\algnum \textbf{end}\;

\algnum \(\hat q_{1-\alpha} \leftarrow \textsc{ConformalQuantile}(\{e_j\}_{j=1}^{m},\alpha)\)\;

\algnum \textbf{for} test scenario \(x\) \textbf{do}\;

\algnum \hspace*{1.5em}\(\hat r(x) \leftarrow f_\theta(x)\)\;

\algnum \hspace*{1.5em}\(C_\alpha(x) \leftarrow [f_\theta(x)-\hat q_{1-\alpha},\, f_\theta(x)+\hat q_{1-\alpha}]\)\;

\algnum \textbf{end}\;

\algnum \textbf{return} \(f_\theta(x), C_\alpha(x)\)\;
\end{minipage}%
\hfill
\begin{minipage}[t]{0.36\linewidth}
\vspace{0pt}
\fbox{%
\begin{minipage}{0.92\linewidth}
\footnotesize
\textbf{ConformalQuantile}

\vspace{0.5mm}
\textbf{Input:} scores \(\{e_j\}_{j=1}^{m}\), miscoverage \(\alpha\).

\vspace{0.5mm}
Sort scores:
\[
e_{(1)}\leq \cdots \leq e_{(m)}.
\]

Set
\[
h=\left\lceil (m+1)(1-\alpha)\right\rceil.
\]

Return
\[
\hat q_{1-\alpha}=e_{(\min\{h,m\})}.
\]

\vspace{0.5mm}
For discrete metrics, replace \(e_j\) with the corresponding classification conformal score.
\end{minipage}
}
\end{minipage}

\end{algorithm}
\end{nolinenumbers}

Algorithm~\ref{app:algorithm} summarizes the full \method{} pipeline. 
Lines 1--4 train the correction model \(g_\phi\) on paired samples and use it to relabel each surrogate-only example with a corrected target \(\tilde r_k\). 
Lines 5--7 merge the trusted paired labels with the corrected surrogate labels to form the augmented training set. 
Line 8 trains the scenario-level evaluator \(f_\theta\) using this augmented set, so that the final predictor depends only on scenario context \(x\). 
Lines 9--12 compute calibration residuals on the held-out calibration set \(\mathcal{D}_c\). 
Line 13 converts these residuals into the split-conformal quantile \(\hat q_{1-\alpha}\), using the finite-sample correction shown in the ConformalQuantile box. 
Lines 14--17 apply the calibrated evaluator to each test scenario and return both the point prediction \(f_\theta(x)\) and its prediction interval \(C_\alpha(x)\). 
For binary metrics, the residual score is replaced by the corresponding classification conformity score, yielding a conformal prediction set instead of a regression interval.

\subsection{nuPlan Dataset Statistics}
\label{app:nuplan_dataset_stats}

\paragraph{Data collection.}
We first train the three ML-based motion planner provided by nuPlan: \texttt{urban\_driver\_open\_loop\_model}, \texttt{vector\_model}, and \texttt{simple\_vector\_model}. These models serve as the target policies to be evaluated. To maintain diversity, each planner dataset covers 62 scenario types, such as crosswalk traversal, traffic-light stopping, and lane changing, from four map locations: Singapore, Boston, Las Vegas, and Pittsburgh. In total, for each planner dataset, we collect approximately \(83.6\)K paired training samples for \(\mathcal{D}_p\), \(15.5\)K paired validation samples for \(\mathcal{D}_v\), \(28.3\)K paired test samples for \(\mathcal{D}_{\mathrm{test}}\), \(1.15\)K held-out calibration samples for \(\mathcal{D}_c\), and \(119\)K--\(135\)K surrogate-only samples for \(\mathcal{D}_s\). Detailed dataset composition is provided below. 

\paragraph{Frozen Urban Driver feature representation.}
For the nuPlan experiments, we use a separately trained Urban Driver feature
encoder to convert each scenario observation into a fixed 256-dimensional
feature vector $x \in \mathbb{R}^{256}$. The encoder is kept frozen during
SCAPE training, and all downstream evaluator models operate only on this
extracted vector $x$.

The input to the encoder contains ego-vehicle history, nearby-agent history,
local vector-map context, and traffic-light state. The ego history is represented
as a temporal polyline whose per-timestep feature is
$(x,y,\theta)\in\mathbb{R}^{3}$. Each nearby-agent polyline uses a
7-dimensional per-timestep feature containing position, heading, velocity, yaw
rate, length, and width:
$
(x,y,\theta,v,\dot{\theta},l,w)\in\mathbb{R}^{7}.
$
Traffic-light status is encoded as a 4-dimensional one-hot vector indicating
green, yellow, red, or unknown state. Local map elements, including lanes, lane
boundaries, stop lines, crosswalks, and route lanes, are represented as vector
polylines in the local ego-centric coordinate frame.

The encoder first maps each raw polyline element to a learned local descriptor
in $\mathbb{R}^{256}$. Thus ego, agent, and map polylines are all embedded into
the same 256-dimensional descriptor space after local encoding, even though
their raw input channel dimensions differ. A global attention module then uses
the ego query to aggregate ego, agent, map, and traffic-light context into a
single scene-level feature vector
$
x \in \mathbb{R}^{256}.
$
This 256-dimensional vector is the only feature used by the SCAPE evaluator and
the neural baselines; the raw ego, agent, map, and traffic-light inputs are not
updated or re-encoded during evaluator training.

\begin{table}[H]
\centering
\small
\caption{Detailed city-level composition of the \method{} nuPlan self-driving experiments dataset splits.}
\label{tab:dataset-city-splits}
\begin{tabular}{llrrrr}
\toprule
Planner & Split & Singapore & Boston & Las Vegas & Pittsburgh \\
\midrule
Urban Driver Open-Loop & Train & 19,504 & 20,145 & 23,833 & 20,123 \\
 & Val & 3,555 & 3,574 & 4,547 & 3,791 \\
 & Test & 6,993 & 6,347 & 7,737 & 7,188 \\
 & Calibration & 264 & 254 & 355 & 276 \\
 & Surrogate & 31,521 & 30,939 & 39,890 & 16,841 \\
\midrule
Vector Model & Train & 19,504 & 20,145 & 23,833 & 20,123 \\
 & Val & 3,555 & 3,574 & 4,547 & 3,791 \\
 & Test & 6,993 & 6,347 & 7,737 & 7,188 \\
 & Calibration & 266 & 254 & 355 & 276 \\
 & Surrogate & 31,532 & 30,939 & 39,890 & 33,118 \\
\midrule
Simple Vector Model & Train & 19,504 & 20,145 & 23,833 & 20,122 \\
 & Val & 3,555 & 3,574 & 4,547 & 3,791 \\
 & Test & 6,993 & 6,347 & 7,737 & 7,188 \\
 & Calibration & 265 & 254 & 355 & 276 \\
 & Surrogate & 31,531 & 30,939 & 39,890 & 33,118 \\
\bottomrule
\end{tabular}
\end{table}

\subsection{Quadruped Velocity-Tracking Dataset Statistics}
\label{app:locomotion_dataset_stats}
\paragraph{Data collection.}
We evaluate a quadruped velocity-tracking policy under two regimes:
\emph{Sim2Sim}, where both the target \(R\) and surrogate \(S\)
come from simulation with different domain-randomization seeds, and
\emph{Sim2Real}, where \(R\) is measured on the physical robot and
\(S\) is obtained from a matched IsaacLab rollout. Each terrain is
reconstructed from an iPhone \texttt{Polycam} scan, yielding a
textured mesh. For Sim2Sim, paired samples \(\mathcal{D}_p\)
span \(4\) indoor/outdoor scenes; for Sim2Real, samples are collected
over \(95\) HLoc~\cite{sarlin2019coarse}-localized deploy poses across \(5\) scenes
(\texttt{indoor}, \texttt{grass}, \texttt{dirt}, \texttt{dirt1},
\texttt{brick}). In total, for the Sim2Sim setting we collect
\(300\) paired training samples for \(\mathcal{D}_p\), \(200\) paired
validation samples for \(\mathcal{D}_v\) that also serve as the held-out calibration set
\(\mathcal{D}_c\), \(200\) paired test samples for \(\mathcal{D}_{\mathrm{test}}\), and \(1{,}493\)
surrogate-only samples for \(\mathcal{D}_s\); for Sim2Real,
\(65/15/15\) paired train/val/test samples and
\(195\) surrogate-only samples. Paired real and reconstructed simulation is shown in Figure~\ref{fig:reconstructed_scene}. Detailed dataset composition is
provided below.
\begin{figure}[H]
    \centering
    \includegraphics[
        width=0.98\linewidth,
        height=0.4\textheight,
        keepaspectratio
    ]{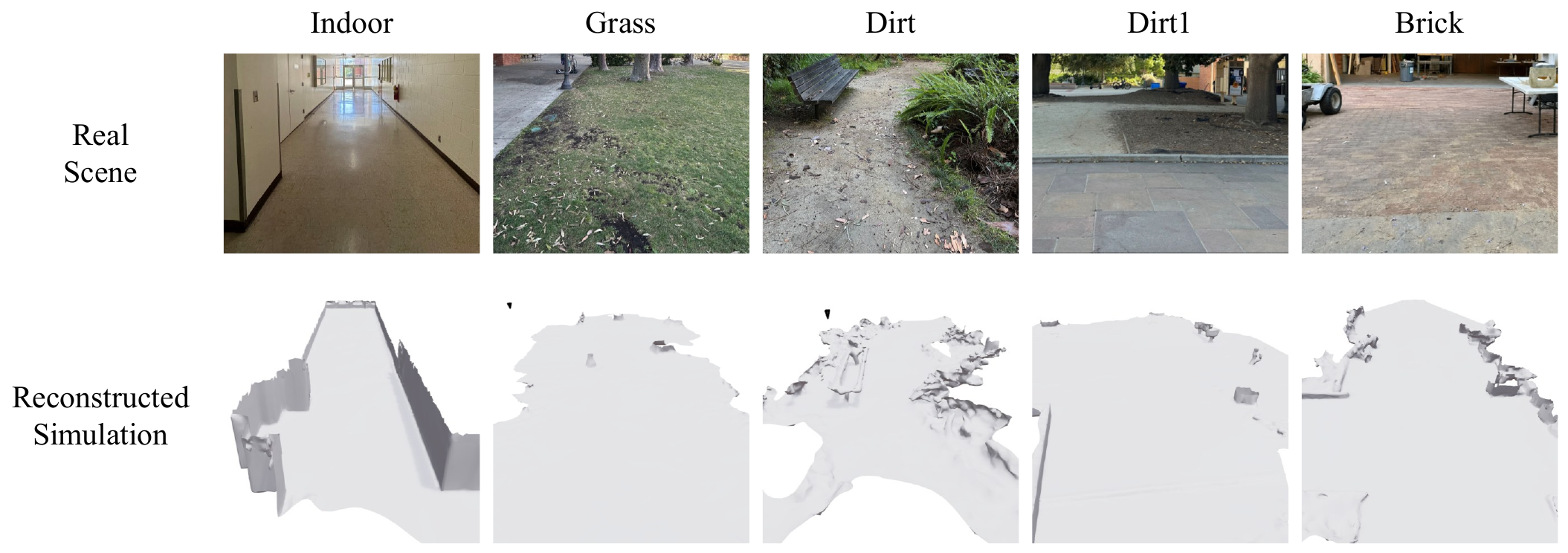}
    \vspace{-2mm}
    \caption{\textbf{Paired real and reconstructed simulation scenes.}
The top row shows real-world scenes, and the bottom row shows their corresponding reconstructed simulation environments.}
\label{fig:reconstructed_scene}
    \vspace{-5mm}
\end{figure}
\begin{table}[H]
\centering
\small
\caption{\textbf{Detailed scene-level composition of the \method{} quadruped velocity tracking dataset splits.} For each setting, the paired pool (\(\mathcal{D}_p\))
is randomly shuffled and split into Train/Val/Test (\(300/200/200\)
for Sim2Sim, \(65/15/15\) for Sim2Real) independently for each of the
\(10\) random seeds, so we report the per-scene totals; surrogate
samples (\(\mathcal{D}_s\)) are unsplit.}
\label{tab:loco-dataset-scene-splits}
\setlength{\tabcolsep}{6pt}
\begin{tabular}{lrrrrrr}
\toprule
\multicolumn{7}{c}{\emph{Sim2Sim}} \\
\midrule
& scene1 & scene2 & scene3 & scene4 & & \textbf{Total} \\
\midrule
Paired (\(\mathcal{D}_p\))    &  59 & 293 &  29 & 319 & & 700 \\
Surrogate (\(\mathcal{D}_s\)) & 109 & 619 &  69 & 696 & & 1{,}493 \\
\midrule
\multicolumn{7}{c}{} \\
\vspace{-2.0mm}\\
\multicolumn{7}{c}{\emph{Sim2Real}} \\
\midrule
& indoor & grass & dirt & dirt1 & brick & \textbf{Total} \\
\midrule
Paired (\(\mathcal{D}_p\))    & 15 & 25 & 25 &  9 & 21 &  95 \\
Surrogate (\(\mathcal{D}_s\)) & 37 & 50 & 46 & 22 & 40 & 195 \\
\bottomrule
\end{tabular}
\end{table}
\paragraph{Frozen quadruped velocity-tracking feature representation.}
For the quadruped velocity-tracking experiments, we represent each locomotion
scenario using a hand-engineered, fixed feature vector
\(x\in\mathbb{R}^{d}\) constructed from the robot's local terrain geometry and
velocity command. Unlike the nuPlan experiments, no learned backbone encoder is
used. This feature vector is kept fixed during SCAPE training, and all downstream
models, including the scenario evaluator \(f_\theta\) and the \method{}
correction model \(g_\phi\), operate only on \(x\).

The terrain component is computed by a body-frame raycaster on the Polycam
terrain mesh and stores per-cell terrain heights relative to the robot base. In
the Sim2Sim setting, we use a \(10\times 8\) local elevation crop covering
\(X\in[-0.25,1.25]\) m and \(Y\in[-0.60,0.60]\) m at 0.15 m resolution, yielding
an 80-dimensional height feature. This is concatenated with the planar command
\([v_x,\omega_z]\), where \(v_x\in[0.1,0.5]\) m s\(^{-1}\) and
\(\omega_z\in[-0.5,0.5]\) rad s\(^{-1}\). We do not include a scene identifier
in Sim2Sim, since the same Go2 policy is evaluated across all four scenes and
both \(R\) and \(S\) are obtained from simulator rollouts. The resulting feature
is \(x\in\mathbb{R}^{82}\), formed by the 80-dimensional elevation crop and the
2-dimensional command.

In the Sim2Real setting, we use a smaller \(11\times 4\) local elevation crop
covering \(X\in[-0.25,1.40]\) m and \(Y\in[-0.30,0.30]\) m at 0.15 m resolution,
which matches the physical raycast used by the deployed policy at run time. The
44-dimensional crop is concatenated with the recorded planar command at each
HLoc-localized deployment pose, where \(v_x\in[0.4,0.6]\) m s\(^{-1}\) and
\(\omega_z\in[-0.2,0.2]\) rad s\(^{-1}\). We additionally include a
5-dimensional scene one-hot vector over \texttt{indoor}, \texttt{grass},
\texttt{dirt}, \texttt{dirt1}, and \texttt{brick}, since Sim2Real residuals are
correlated with visually and tactilely different real-world terrains. The
resulting feature is \(x\in\mathbb{R}^{51}\), formed by the 44-dimensional
elevation crop, the 2-dimensional command, and the 5-dimensional scene
identifier.

\subsection{Baseline Definition}
\label{app:scape_baselines}
\begin{table}[t]
\centering
\small
\captionsetup{skip=3pt}
\vspace{-6pt}
\caption{\textbf{Baseline definitions.} Statistical baselines estimate aggregate target performance, while neural baselines train scenario-level predictors \(f_\theta(x)\).}
\label{tab:baseline_definitions}
\setlength{\tabcolsep}{6pt}
\renewcommand{\arraystretch}{1.35}
\begin{tabular}{llp{0.75\linewidth}}
\toprule
\textbf{Type} & \textbf{Method} & \textbf{Definition} \\
\midrule
Statistical & MC 
& \(\displaystyle \hat\mu_{\mathrm{MC}}=\frac{1}{n}\sum_i r_i\) \\

Statistical & CV 
& \(\displaystyle 
\hat\mu_{\mathrm{CV}}
=
\frac{1}{n}\sum_i(r_i-\hat\beta s_i)
+
\frac{1}{\ell}\sum_k\hat\beta s_k,\quad
\hat\beta=
\frac{\ell}{n+\ell}
\frac{\widehat{\mathrm{Cov}}(s,r)}
{\widehat{\mathrm{Var}}(s)}
\) \\

Statistical & PPI 
& \(\displaystyle 
\hat\mu_{\mathrm{PPI}}
=
\frac{1}{\ell}\sum_k s_k
+
\frac{1}{n}\sum_i(r_i-s_i)
\) \\

\midrule
Neural & R-Only 
& \(\displaystyle 
\mathcal{L}_{\mathrm{R}}
=
\frac{1}{n}\sum_i
\mathcal{L}(f_\theta(x_i),r_i)
\) \\

Neural & RS-Mix 
& \(\displaystyle 
\mathcal{L}_{\mathrm{RS}}
=
\mathcal{L}_{\mathrm{R}}
+
\frac{1}{\ell}\sum_k
\mathcal{L}(f_\theta(x_k),s_k)
\) \\

Neural & PPI-NN 
& \(\displaystyle 
\mathcal{L}_{\mathrm{PPI\text{-}NN}}
=
\frac{1}{\ell}\sum_k
\mathcal{L}(f_\theta(x_k),s_k)
+
\frac{1}{n}\sum_i
\big[
\mathcal{L}(f_\theta(x_i),r_i)
-
\mathcal{L}(f_\theta(x_i),s_i)
\big]
\) \\
\bottomrule
\end{tabular}
\end{table}

Table~\ref{tab:baseline_definitions} summarizes the baselines used in our experiments. 
The first group contains statistical aggregate estimators, which do not condition on scenario observations and therefore provide a single scenario-independent target estimate. 
MC estimates target performance using only paired target labels, serving as the real-data-only reference. 
CV uses the surrogate metric as a control variate to reduce the variance of the target mean estimate when \(S\) and \(R\) are correlated. 
PPI estimates the surrogate mean on the large simulation-only set and corrects its bias using paired residuals, representing a prediction-powered aggregate estimator.

The second group contains scenario-conditioned neural predictors \(f_\theta(x)\), which are directly comparable to \method{} in producing per-scenario predictions. 
R-Only trains only on paired target labels and measures the performance achievable without using surrogate-only data. 
RS-Mix naively augments the paired target labels with raw surrogate labels, testing whether low-cost simulation data can be used directly despite sim-to-real bias. 
PPI-NN adapts the PPI correction objective to neural training by combining surrogate-only supervision with a paired residual correction term, testing correction during training as an alternative to the correction-before-augmentation design of \method{}.

\subsection{\method{} MLP Hyperparameters}
\label{app:nuplan_mlp_hyperparameters}

\paragraph{Training objectives and label weighting.}
The correction model $g_\phi$ is trained on the paired data:
\[
\mathcal{L}_{g}
=
\frac{1}{n}
\sum_{i=1}^{n}
\mathcal{L}\bigl(g_\phi(x_i,s_i),r_i\bigr),
\qquad
\tilde r_k=g_\phi(x_k,s_k).
\]
The final predictor $f_\theta$ is then trained on the real and corrected pseudo labels:
\[
\mathcal{L}_{f}
=
\frac{1}{n+\ell}
\left(
\sum_{i=1}^{n}
\mathcal{L}\bigl(f_\theta(x_i),r_i\bigr)
+
w_{\mathrm{pseudo}}
\sum_{k=1}^{\ell}
\mathcal{L}\bigl(f_\theta(x_k),\tilde r_k\bigr)
\right).
\]
All main experiments use $w_{\mathrm{pseudo}}=1$, assigning equal per-sample weights to real and corrected pseudo labels without dataset-specific tuning.

\begin{table}[t]
\centering
\small
\vspace{-6pt}
\captionsetup{skip=3pt}
\caption{\textbf{MLP architecture and training hyperparameters (nuPlan).}
The scenario evaluator \(f_\theta\) is used by R-Only, RS-Mix, PPI-NN, and the final evaluator in \method{}. The correction model \(g_\phi\) additionally takes the surrogate metric vector \(s\) as input and predicts corrected target metrics \(\tilde r\). The four output dimensions correspond to ADE, TTC \(<1\)s, drivable-area compliance, and no ego at-fault collision.}
\label{tab:mlp-hyperparameters}
\setlength{\tabcolsep}{4pt}
\begin{tabular}{lcc}
\toprule
\textbf{Component} & \textbf{Scenario evaluator \(f_\theta\)} & \textbf{Correction model \(g_\phi\)} \\
\midrule
Input & \(x \in \mathbb{R}^{256}\) & \([x,s] \in \mathbb{R}^{260}\) \\
Output & \(\hat r=f_\theta(x)\in\mathbb{R}^{4}\) & \(\tilde r=g_\phi(x,s)\in\mathbb{R}^{4}\) \\
Hidden layers & \([1024,512,256,128]\) & \([1024,512,256,128]\) \\
Activation & ReLU & ReLU \\
Dropout & \(0.2\) & \(0.2\) \\
Trainable parameters & \(0.953\)M & \(0.957\)M \\
Loss & MSE \(+\) 3 BCE & MSE \(+\) 3 BCE \\
Optimizer & Adam & Adam \\
Learning rate & \(3\times 10^{-5}\) & \(3\times 10^{-5}\) \\
Weight decay & \(1\times 10^{-4}\) & \(1\times 10^{-4}\) \\
Batch size & \(256\) & \(256\) \\
Epochs & \(500\) & \(500\) \\
\bottomrule
\end{tabular}
\end{table}

{\bf nuPlan}. For both \(f_\theta\) and \(g_\phi\), all output metrics are weighted equally. 
We use mean squared error (MSE) for the continuous average displacement error (ADE) metric and binary cross-entropy (BCE) for the binary metrics, including time-to-collision (TTC) under \(1\)s, drivable-area compliance (DA), and no ego at-fault collision (NAFC):
\vspace{2pt}
\[
\mathcal{L}(\hat r,r)
=
\mathrm{MSE}_{\mathrm{ADE}}
+
\mathrm{BCE}_{\mathrm{TTC}}
+
\mathrm{BCE}_{\mathrm{DA}}
+
\mathrm{BCE}_{\mathrm{NAFC}}.
\]

{\bf Quadruped.} For both \(f_\theta\) and \(g_\phi\), the two output metrics are weighted
equally. Given prediction \(\hat r=(\hat v,\hat \omega)\) and target
\(r=(v,\omega)\), we use
\[
\mathcal{L}(\hat r, r)
=
\mathrm{MSE}_{v}(\hat v, v)
+
\mathrm{MSE}_{\omega}(\hat \omega, \omega),
\]

where \(v\) and \(\omega\) are the per-step velocity- and yaw-rate-tracking
errors (m/s and rad/s) computed over a \(2\)-s rollout. MSE targets are
z-scored using the training-split statistics.

\begin{table}[t]
\centering
\small
\vspace{-5pt}
\captionsetup{skip=3pt}
\caption{\textbf{MLP architecture and training hyperparameters (quadruped).}
The scenario evaluator \(f_\theta\) is used by R-Only, RS-Mix, PPI-NN, and the
final evaluator in \method{}. The correction model \(g_\phi\) additionally
takes the surrogate metric vector \(s\) as input and predicts corrected
target metrics \(\tilde r\). The two output dimensions correspond to the
per-step velocity tracking error \(v\) and yaw-rate tracking error \(\omega\).
Input dimensionality \(d{=}51\) (\(44\)-D \(11{\times}4\) ground-height crop
\(+\) \([v_x, \omega_z]\) command \(+\) \(5\)-D scene one-hot) for the
sim-to-real experiment, and \(d{=}82\) for the in-sim experiment.}
\label{tab:mlp-hyperparameters-loco}
\setlength{\tabcolsep}{4pt}
\begin{tabular}{lcc}
\toprule
\textbf{Component} & \textbf{Scenario evaluator \(f_\theta\)} & \textbf{Correction model \(g_\phi\)} \\
\midrule
Input & \(x \in \mathbb{R}^{d}\) & \([x,s] \in \mathbb{R}^{d+2}\) \\
Output & \(\hat r=f_\theta(x)\in\mathbb{R}^{2}\) & \(\tilde r=g_\phi(x,s)\in\mathbb{R}^{2}\) \\
Hidden layers & \([128, 64, 32]\) & \([128, 64, 32]\) \\
Activation & ReLU & ReLU \\
Dropout & \(0.3\) & \(0.3\) \\
Trainable parameters & \(16.8\)K & \(17.1\)K \\
Loss & 2 MSE & 2 MSE \\
Target normalization & z-score (train split) & z-score (train split) \\
Optimizer & Adam & Adam \\
Learning rate & \(3\times 10^{-4}\) & \(3\times 10^{-4}\) \\
Weight decay & \(1\times 10^{-3}\) & \(1\times 10^{-3}\) \\
Batch size & \(64\) & \(64\) \\
Epochs & \(500\) & \(500\) \\
\bottomrule
\end{tabular}
\end{table}
\vspace{-6pt}

\subsection{Evaluation Metrics and Reported Quantities}
\label{app:experiments_metrics}
We summarize how the reported quantities are computed. 
For a fixed target metric, let \(\mathcal{T}=\{(x_i,r_i)\}_{i=1}^{n}\) denote the test set, where \(x_i\) is a scenario and \(r_i\) is the target measurement. 
For \method{}, let \(\hat r_i\) denote its prediction on \(x_i\); for a baseline \(B\), let \(\hat r_i^{B}\) denote the corresponding baseline prediction. 
For binary metrics, the prediction is the positive-class probability.

\paragraph{Task performance metrics.}
For nuPlan, we evaluate each planner using four target metrics. Average
displacement error (ADE) is the mean displacement between the planned ego
trajectory and the logged expert trajectory. TTC \(<1\)s is a binary safety
metric indicating whether the ego vehicle reaches a time-to-collision below one
second. Drivable-area compliance and no ego at-fault collision are binary
metrics indicating whether the ego trajectory stays within the drivable area and
avoids at-fault collisions, respectively. For quadruped velocity tracking, each target measurement is a rollout-level
tracking error computed over a 2-s rollout. \(\mathrm{MAE}_{\mathrm{vel}}\) and
\(\mathrm{MAE}_{\mathrm{yaw}}\) denote the mean absolute errors between the
commanded and measured forward velocity and yaw rate, respectively. Lower values
indicate better tracking.

\paragraph{Test loss.}
For nuPlan continuous targets, we report mean squared error:
\begin{equation}
\mathcal{L}(\method{})
=
\frac{1}{n}
\sum_{i=1}^{n}
\left(
\hat r_i-r_i
\right)^2 .
\end{equation}
For nuPlan binary targets, we report binary cross entropy:
\begin{equation}
\mathcal{L}(\method{})
=
-\frac{1}{n}
\sum_{i=1}^{n}
\left[
r_i\log \hat r_i
+
(1-r_i)\log(1-\hat r_i)
\right].
\end{equation}

For quadruped continuous velocity tracking targets, we report mean absolute prediction error on the
rollout-level tracking-error targets:
\begin{equation}
\mathcal{L}(\method{})
=
\frac{1}{n}
\sum_{i=1}^{n}
\left|
\hat r_i-r_i
\right| .
\end{equation}

The same metric-specific definition is used for each baseline \(B\) by replacing
\(\hat r_i\) with \(\hat r_i^B\). All prediction-loss results, including the
main paired-data sweep, OOD experiments, and surrogate-scaling experiments, are
computed using the corresponding task-specific test loss.

\paragraph{Relative loss reduction.}
Given a baseline \(B\), the relative test-loss reduction of \method{} is
\begin{equation}
\Delta_{\mathrm{loss}}(\method{},B)
=
100\%
\left(
\frac{
\mathcal{L}(B)-\mathcal{L}(\method{})
}{
\mathcal{L}(B)
}
\right).
\end{equation}
When multiple metrics, planners, paired-data fractions, or random seeds are involved, we first compute the loss for each setting and then report the average relative reduction over the corresponding collection of settings.

\paragraph{Paired-label efficiency.}
Let \(\rho_p\in(0,1]\) denote the fraction of the full paired training set \(\mathcal{D}_p\) used for training. 
For a target loss level \(\tau\), let \(\rho_{\method{}}(\tau)\) be the smallest paired-data fraction at which \method{} reaches loss no larger than \(\tau\), and define \(\rho_B(\tau)\) analogously for baseline \(B\). 
The paired-label saving of \method{} relative to \(B\) is
\begin{equation}
\mathrm{LabelEfficiency}(\method{},B;\tau)
=
100\%
\left(
1-
\frac{\rho_{\method{}}(\tau)}{\rho_B(\tau)}
\right).
\end{equation}
In the main results, \(\tau\) is chosen as the strongest full-budget baseline performance, and we report how much paired data \method{} saves while matching that target performance.

\paragraph{Surrogate-data scaling capacity.}
Let \(\rho_s\in(0,1]\) denote the fraction of the simulation-only set \(\mathcal{D}_s\) used for training. 
To study correction capacity, we fix \(\rho_p\), vary \(\rho_s\), retrain each scenario-conditioned method, and report the resulting test loss \(\mathcal{L}(\rho_p,\rho_s)\). 
This measures how much surrogate-only data can be safely converted into useful supervision under a fixed paired-label budget. 
For each paired-data fraction, the best surrogate fraction is the \(\rho_s\) that gives the lowest test loss.

\paragraph{nuPlan OOD evaluation.}
The nuPlan OOD experiment uses the same test-loss definition above. 
For OOD evaluation, paired target labels from one city are removed during training, while evaluation is performed on the original test set for direct comparison with the standard full-training setting. 
We compare training with and without OOD surrogate-only data by the relative loss reduction:
\begin{equation}
\Delta_{\mathrm{OOD\text{-}S}}(\method{})
=
100\%
\left(
\frac{
\mathcal{L}_{\mathrm{no\text{-}OOD\text{-}S}}(\method{})
-
\mathcal{L}_{\mathrm{with\text{-}OOD\text{-}S}}(\method{})
}{
\mathcal{L}_{\mathrm{no\text{-}OOD\text{-}S}}(\method{})
}
\right).
\end{equation}
\paragraph{Conformal calibration.}
For conformal uncertainty evaluation, each trained evaluator is calibrated on
the held-out calibration set
\(\mathcal{D}_c=\{(x_j,r_j)\}_{j=1}^{m}\). 
For continuous metrics, we compute residual scores $e_j = |r_j-\hat r_j|$. Given miscoverage level \(\alpha\), the split-conformal threshold is
\begin{equation}
\hat q_{1-\alpha}
=
\mathrm{Quantile}_{\lceil (m+1)(1-\alpha)\rceil/m}
\left(
\{e_j\}_{j=1}^{m}
\right),
\end{equation}
and the prediction interval is
\begin{equation}
C_\alpha(x)
=
[
\hat r(x)-\hat q_{1-\alpha},
\hat r(x)+\hat q_{1-\alpha}
].
\end{equation}
For binary metrics, we compute the corresponding classification conformity
scores and return a conformal prediction set
\(C_\alpha(x)\subseteq\{0,1\}\).

For nuPlan, the continuous-metric procedure is used for ADE, while the binary
conformal procedure is used for TTC \(<1\)s, drivable-area compliance, and no
ego at-fault collision. For quadruped, both
\(\mathrm{MAE}_{\mathrm{vel}}\) and \(\mathrm{MAE}_{\mathrm{yaw}}\) are continuous metrics and use the same residual-based conformal interval.

We verify empirical coverage on the test set:
\begin{equation}
\widehat{\mathrm{Cov}}
=
\frac{1}{n}
\sum_{i=1}^{n}
\mathbf{1}
\left[
r_i \in C_\alpha(x_i)
\right].
\end{equation}
For continuous metrics, the reported conformal width is the interval length
\(2\hat q_{1-\alpha}\). For binary metrics, the reported width is the average
prediction-set size:
\begin{equation}
W
=
\frac{1}{n}
\sum_{i=1}^{n}
|C_\alpha(x_i)|.
\end{equation}

\subsection{Additional nuPlan Scenario-Level Prediction Results}
\label{app:nuplan_main_results}
\begin{figure}[H]
    \centering
    \vspace{0pt}
    \captionsetup{skip=3pt}

    \begin{subfigure}[t]{0.48\linewidth}
        \centering
        \includegraphics[width=\linewidth]{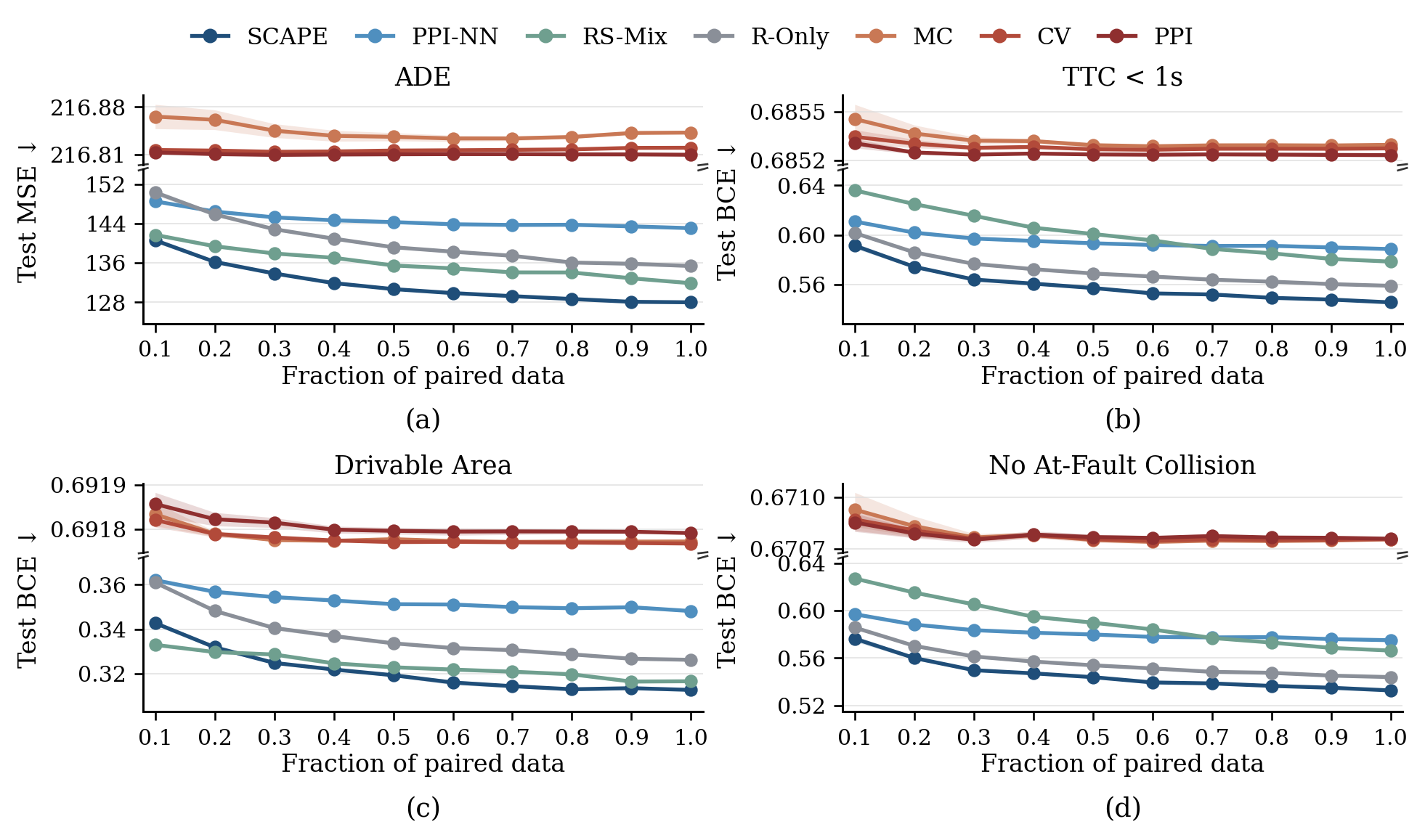}
        \caption{Vector}
        \label{fig:appendix-scenario-pred-vector}
    \end{subfigure}
    \hfill
    \begin{subfigure}[t]{0.48\linewidth}
        \centering
        \includegraphics[width=\linewidth]{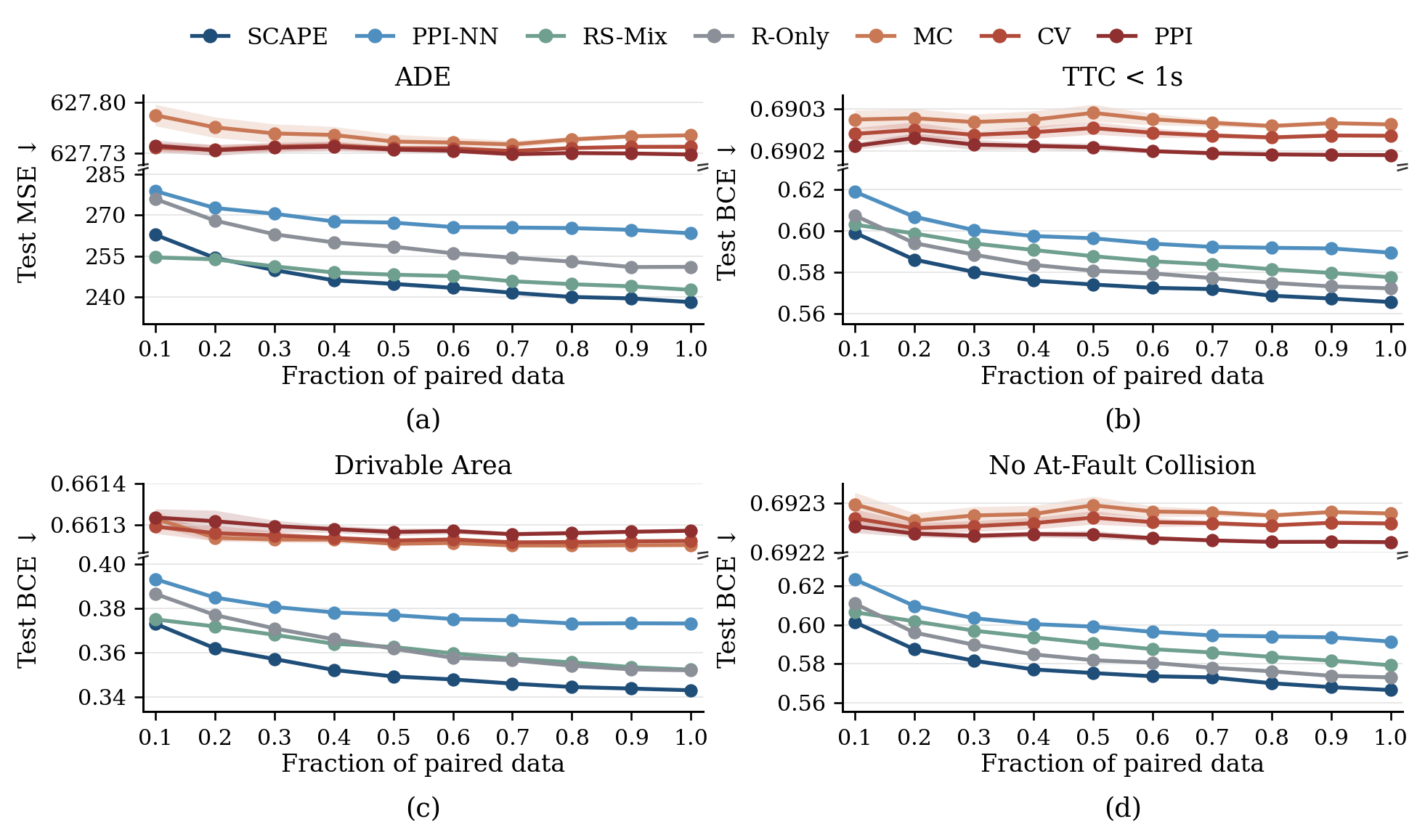}
        \caption{Simple Vector}
        \label{fig:appendix-scenario-pred-simplevector}
    \end{subfigure}

    \caption{\small \textbf{Additional scenario-level prediction results on nuPlan.}
Panels (a) and (b) report Vector and Simple Vector planner results, respectively.
Test loss is reported as the paired-data fraction varies; curves show mean \(\pm\) standard error over 10 random paired-data seeds.}
    \label{fig:appendix-scenario-pred-nuplan}
    \vspace{-8pt}
\end{figure}

\subsection{Full Conformal Prediction Results}
\label{app:conformal full}
\newcommand{\confcompactsetup}{%
    \captionsetup{skip=7pt}
    \scriptsize
    \setlength{\tabcolsep}{2.0pt}
    \renewcommand{\arraystretch}{1.02}
}
\newcommand{\confblockcaption}[5]{%
\caption{\small \textbf{#1}
Columns are paired-data fractions. Cells are mean~$\pm$~standard error over 10 paired-data seeds; bold marks the narrowest method per column. #2 \emph{cov.} reports empirical test coverage range across neural methods.}
\label{#3}
\confcompactsetup
\noindent\textbf{#4.} \emph{#5}
\vspace{0pt}
}

\newcommand{\confsubtabletitle}[2]{%
\vspace{3pt}
\noindent\textbf{#1.} \emph{#2}
\vspace{0pt}
}

\begin{table}[H]
\centering
\vspace{-3pt}
\confblockcaption
{Full split conformal width sweep on nuPlan Urban Driver.}
{For ADE, width is interval length in meters; for binary targets, width is average prediction-set size.}
{tab:conformal-urban-full}
{ADE, interval width (m)}
{cov.\ 0.946--0.953}
\resizebox{\textwidth}{!}{%
\begin{tabular}{l|cccccccccc}
\toprule
Method & .1 & .2 & .3 & .4 & .5 & .6 & .7 & .8 & .9 & 1.0 \\
\midrule
\method{} & $\boldsymbol{44.92\!\pm\!0.26}$ & $\boldsymbol{43.23\!\pm\!0.23}$ & $\boldsymbol{42.10\!\pm\!0.34}$ & $\boldsymbol{41.69\!\pm\!0.22}$ & $\boldsymbol{41.00\!\pm\!0.35}$ & $\boldsymbol{40.64\!\pm\!0.26}$ & $\boldsymbol{40.63\!\pm\!0.26}$ & $\boldsymbol{40.35\!\pm\!0.26}$ & $\boldsymbol{39.93\!\pm\!0.26}$ & $\boldsymbol{40.07\!\pm\!0.26}$ \\
PPI-NN & $45.21\!\pm\!0.26$ & $45.11\!\pm\!0.21$ & $44.91\!\pm\!0.32$ & $44.76\!\pm\!0.18$ & $44.60\!\pm\!0.09$ & $44.25\!\pm\!0.16$ & $44.55\!\pm\!0.18$ & $44.44\!\pm\!0.14$ & $44.17\!\pm\!0.21$ & $44.44\!\pm\!0.16$ \\
RS-Mix & $49.09\!\pm\!0.21$ & $48.52\!\pm\!0.18$ & $47.90\!\pm\!0.25$ & $46.78\!\pm\!0.30$ & $47.08\!\pm\!0.30$ & $46.38\!\pm\!0.31$ & $45.41\!\pm\!0.12$ & $45.66\!\pm\!0.20$ & $45.45\!\pm\!0.18$ & $44.65\!\pm\!0.16$ \\
R-Only & $46.11\!\pm\!0.22$ & $44.73\!\pm\!0.35$ & $43.82\!\pm\!0.18$ & $43.68\!\pm\!0.31$ & $42.80\!\pm\!0.30$ & $42.27\!\pm\!0.22$ & $42.00\!\pm\!0.22$ & $41.98\!\pm\!0.21$ & $40.99\!\pm\!0.16$ & $41.02\!\pm\!0.30$ \\
MC & $50.21\!\pm\!0.11$ & $50.22\!\pm\!0.09$ & $50.26\!\pm\!0.07$ & $50.24\!\pm\!0.04$ & $50.25\!\pm\!0.03$ & $50.28\!\pm\!0.03$ & $50.28\!\pm\!0.02$ & $50.28\!\pm\!0.02$ & $50.28\!\pm\!0.01$ & $50.27\!\pm\!0.00$ \\
CV & $50.37\!\pm\!0.04$ & $50.32\!\pm\!0.03$ & $50.31\!\pm\!0.02$ & $50.29\!\pm\!0.02$ & $50.29\!\pm\!0.02$ & $50.29\!\pm\!0.01$ & $50.29\!\pm\!0.01$ & $50.29\!\pm\!0.01$ & $50.29\!\pm\!0.00$ & $50.29\!\pm\!0.00$ \\
PPI & $50.41\!\pm\!0.05$ & $50.35\!\pm\!0.03$ & $50.33\!\pm\!0.03$ & $50.32\!\pm\!0.03$ & $50.31\!\pm\!0.02$ & $50.31\!\pm\!0.01$ & $50.29\!\pm\!0.01$ & $50.30\!\pm\!0.01$ & $50.31\!\pm\!0.01$ & $50.31\!\pm\!0.00$ \\
\bottomrule
\end{tabular}
}

\confsubtabletitle{TTC$<1$s, average set size}{cov.\ 0.941--0.957}
\resizebox{\textwidth}{!}{%
\begin{tabular}{l|cccccccccc}
\toprule
Method & .1 & .2 & .3 & .4 & .5 & .6 & .7 & .8 & .9 & 1.0 \\
\midrule
\method{} & $\boldsymbol{1.600\!\pm\!0.006}$ & $\boldsymbol{1.558\!\pm\!0.006}$ & $\boldsymbol{1.532\!\pm\!0.007}$ & $\boldsymbol{1.516\!\pm\!0.009}$ & $\boldsymbol{1.492\!\pm\!0.003}$ & $\boldsymbol{1.486\!\pm\!0.007}$ & $\boldsymbol{1.472\!\pm\!0.006}$ & $\boldsymbol{1.468\!\pm\!0.006}$ & $\boldsymbol{1.453\!\pm\!0.005}$ & $\boldsymbol{1.447\!\pm\!0.010}$ \\
PPI-NN & $1.665\!\pm\!0.006$ & $1.643\!\pm\!0.005$ & $1.627\!\pm\!0.004$ & $1.623\!\pm\!0.004$ & $1.620\!\pm\!0.006$ & $1.618\!\pm\!0.004$ & $1.619\!\pm\!0.006$ & $1.614\!\pm\!0.005$ & $1.613\!\pm\!0.005$ & $1.614\!\pm\!0.004$ \\
RS-Mix & $1.654\!\pm\!0.004$ & $1.632\!\pm\!0.004$ & $1.609\!\pm\!0.006$ & $1.591\!\pm\!0.008$ & $1.577\!\pm\!0.005$ & $1.559\!\pm\!0.004$ & $1.540\!\pm\!0.006$ & $1.526\!\pm\!0.006$ & $1.519\!\pm\!0.006$ & $1.505\!\pm\!0.006$ \\
R-Only & $1.623\!\pm\!0.005$ & $1.586\!\pm\!0.007$ & $1.564\!\pm\!0.004$ & $1.555\!\pm\!0.006$ & $1.543\!\pm\!0.006$ & $1.540\!\pm\!0.004$ & $1.536\!\pm\!0.004$ & $1.528\!\pm\!0.005$ & $1.512\!\pm\!0.003$ & $1.497\!\pm\!0.003$ \\
MC & $2.000\!\pm\!0.000$ & $2.000\!\pm\!0.000$ & $2.000\!\pm\!0.000$ & $2.000\!\pm\!0.000$ & $2.000\!\pm\!0.000$ & $2.000\!\pm\!0.000$ & $2.000\!\pm\!0.000$ & $2.000\!\pm\!0.000$ & $2.000\!\pm\!0.000$ & $2.000\!\pm\!0.000$ \\
CV & $2.000\!\pm\!0.000$ & $2.000\!\pm\!0.000$ & $2.000\!\pm\!0.000$ & $2.000\!\pm\!0.000$ & $2.000\!\pm\!0.000$ & $2.000\!\pm\!0.000$ & $2.000\!\pm\!0.000$ & $2.000\!\pm\!0.000$ & $2.000\!\pm\!0.000$ & $2.000\!\pm\!0.000$ \\
PPI & $2.000\!\pm\!0.000$ & $2.000\!\pm\!0.000$ & $2.000\!\pm\!0.000$ & $2.000\!\pm\!0.000$ & $2.000\!\pm\!0.000$ & $2.000\!\pm\!0.000$ & $2.000\!\pm\!0.000$ & $2.000\!\pm\!0.000$ & $2.000\!\pm\!0.000$ & $2.000\!\pm\!0.000$ \\
\bottomrule
\end{tabular}
}

\confsubtabletitle{Drivable Area, average set size}{cov.\ 0.929--0.941}
\resizebox{\textwidth}{!}{%
\begin{tabular}{l|cccccccccc}
\toprule
Method & .1 & .2 & .3 & .4 & .5 & .6 & .7 & .8 & .9 & 1.0 \\
\midrule
\method{} & $1.219\!\pm\!0.006$ & $\boldsymbol{1.173\!\pm\!0.006}$ & $\boldsymbol{1.156\!\pm\!0.001}$ & $\boldsymbol{1.153\!\pm\!0.004}$ & $\boldsymbol{1.135\!\pm\!0.002}$ & $\boldsymbol{1.132\!\pm\!0.003}$ & $\boldsymbol{1.131\!\pm\!0.003}$ & $\boldsymbol{1.126\!\pm\!0.004}$ & $\boldsymbol{1.118\!\pm\!0.003}$ & $\boldsymbol{1.118\!\pm\!0.004}$ \\
PPI-NN & $1.286\!\pm\!0.004$ & $1.263\!\pm\!0.004$ & $1.255\!\pm\!0.003$ & $1.246\!\pm\!0.003$ & $1.245\!\pm\!0.003$ & $1.240\!\pm\!0.003$ & $1.238\!\pm\!0.003$ & $1.235\!\pm\!0.004$ & $1.231\!\pm\!0.003$ & $1.233\!\pm\!0.004$ \\
RS-Mix & $\boldsymbol{1.217\!\pm\!0.004}$ & $1.202\!\pm\!0.004$ & $1.199\!\pm\!0.001$ & $1.186\!\pm\!0.003$ & $1.176\!\pm\!0.004$ & $1.173\!\pm\!0.002$ & $1.174\!\pm\!0.003$ & $1.162\!\pm\!0.003$ & $1.153\!\pm\!0.004$ & $1.152\!\pm\!0.002$ \\
R-Only & $1.273\!\pm\!0.004$ & $1.231\!\pm\!0.005$ & $1.207\!\pm\!0.002$ & $1.193\!\pm\!0.003$ & $1.183\!\pm\!0.004$ & $1.178\!\pm\!0.004$ & $1.166\!\pm\!0.004$ & $1.165\!\pm\!0.004$ & $1.156\!\pm\!0.002$ & $1.155\!\pm\!0.003$ \\
MC & $2.000\!\pm\!0.000$ & $2.000\!\pm\!0.000$ & $2.000\!\pm\!0.000$ & $2.000\!\pm\!0.000$ & $2.000\!\pm\!0.000$ & $2.000\!\pm\!0.000$ & $2.000\!\pm\!0.000$ & $2.000\!\pm\!0.000$ & $2.000\!\pm\!0.000$ & $2.000\!\pm\!0.000$ \\
CV & $2.000\!\pm\!0.000$ & $2.000\!\pm\!0.000$ & $2.000\!\pm\!0.000$ & $2.000\!\pm\!0.000$ & $2.000\!\pm\!0.000$ & $2.000\!\pm\!0.000$ & $2.000\!\pm\!0.000$ & $2.000\!\pm\!0.000$ & $2.000\!\pm\!0.000$ & $2.000\!\pm\!0.000$ \\
PPI & $2.000\!\pm\!0.000$ & $2.000\!\pm\!0.000$ & $2.000\!\pm\!0.000$ & $2.000\!\pm\!0.000$ & $2.000\!\pm\!0.000$ & $2.000\!\pm\!0.000$ & $2.000\!\pm\!0.000$ & $2.000\!\pm\!0.000$ & $2.000\!\pm\!0.000$ & $2.000\!\pm\!0.000$ \\
\bottomrule
\end{tabular}
}

\confsubtabletitle{No At-Fault Collision, average set size}{cov.\ 0.947--0.960}
\resizebox{\textwidth}{!}{%
\begin{tabular}{l|cccccccccc}
\toprule
Method & .1 & .2 & .3 & .4 & .5 & .6 & .7 & .8 & .9 & 1.0 \\
\midrule
\method{} & $\boldsymbol{1.573\!\pm\!0.006}$ & $\boldsymbol{1.532\!\pm\!0.008}$ & $\boldsymbol{1.505\!\pm\!0.005}$ & $\boldsymbol{1.498\!\pm\!0.007}$ & $\boldsymbol{1.469\!\pm\!0.003}$ & $\boldsymbol{1.472\!\pm\!0.007}$ & $\boldsymbol{1.456\!\pm\!0.004}$ & $\boldsymbol{1.452\!\pm\!0.004}$ & $\boldsymbol{1.444\!\pm\!0.006}$ & $\boldsymbol{1.430\!\pm\!0.004}$ \\
PPI-NN & $1.637\!\pm\!0.006$ & $1.616\!\pm\!0.005$ & $1.597\!\pm\!0.005$ & $1.590\!\pm\!0.003$ & $1.586\!\pm\!0.003$ & $1.577\!\pm\!0.005$ & $1.577\!\pm\!0.005$ & $1.570\!\pm\!0.004$ & $1.573\!\pm\!0.005$ & $1.573\!\pm\!0.004$ \\
RS-Mix & $1.608\!\pm\!0.003$ & $1.590\!\pm\!0.004$ & $1.574\!\pm\!0.005$ & $1.557\!\pm\!0.005$ & $1.544\!\pm\!0.004$ & $1.531\!\pm\!0.006$ & $1.520\!\pm\!0.004$ & $1.510\!\pm\!0.005$ & $1.506\!\pm\!0.004$ & $1.495\!\pm\!0.003$ \\
R-Only & $1.597\!\pm\!0.004$ & $1.556\!\pm\!0.006$ & $1.540\!\pm\!0.006$ & $1.529\!\pm\!0.006$ & $1.517\!\pm\!0.004$ & $1.520\!\pm\!0.003$ & $1.509\!\pm\!0.004$ & $1.498\!\pm\!0.007$ & $1.488\!\pm\!0.004$ & $1.474\!\pm\!0.009$ \\
MC & $2.000\!\pm\!0.000$ & $2.000\!\pm\!0.000$ & $2.000\!\pm\!0.000$ & $2.000\!\pm\!0.000$ & $2.000\!\pm\!0.000$ & $2.000\!\pm\!0.000$ & $2.000\!\pm\!0.000$ & $2.000\!\pm\!0.000$ & $2.000\!\pm\!0.000$ & $2.000\!\pm\!0.000$ \\
CV & $2.000\!\pm\!0.000$ & $2.000\!\pm\!0.000$ & $2.000\!\pm\!0.000$ & $2.000\!\pm\!0.000$ & $2.000\!\pm\!0.000$ & $2.000\!\pm\!0.000$ & $2.000\!\pm\!0.000$ & $2.000\!\pm\!0.000$ & $2.000\!\pm\!0.000$ & $2.000\!\pm\!0.000$ \\
PPI & $2.000\!\pm\!0.000$ & $2.000\!\pm\!0.000$ & $2.000\!\pm\!0.000$ & $2.000\!\pm\!0.000$ & $2.000\!\pm\!0.000$ & $2.000\!\pm\!0.000$ & $2.000\!\pm\!0.000$ & $2.000\!\pm\!0.000$ & $2.000\!\pm\!0.000$ & $2.000\!\pm\!0.000$ \\
\bottomrule
\end{tabular}
}
\vspace{-6pt}
\end{table}

\begin{table}[H]
\centering
\vspace{-4pt}
\confblockcaption
{Full split conformal width sweep on nuPlan Vector.}
{For ADE, width is interval length in meters; for binary targets, width is average prediction-set size.}
{tab:conformal-vector-full}
{ADE, interval width (m)}
{cov.\ 0.942--0.955}
\resizebox{\textwidth}{!}{%
\begin{tabular}{l|cccccccccc}
\toprule
Method & .1 & .2 & .3 & .4 & .5 & .6 & .7 & .8 & .9 & 1.0 \\
\midrule
\method{} & $48.78\!\pm\!0.47$ & $\boldsymbol{46.67\!\pm\!0.44}$ & $\boldsymbol{45.88\!\pm\!0.40}$ & $\boldsymbol{44.82\!\pm\!0.25}$ & $\boldsymbol{45.13\!\pm\!0.48}$ & $\boldsymbol{43.64\!\pm\!0.35}$ & $\boldsymbol{43.75\!\pm\!0.43}$ & $\boldsymbol{43.77\!\pm\!0.41}$ & $\boldsymbol{43.27\!\pm\!0.29}$ & $\boldsymbol{43.43\!\pm\!0.29}$ \\
PPI-NN & $51.20\!\pm\!0.29$ & $51.54\!\pm\!0.20$ & $50.72\!\pm\!0.22$ & $50.44\!\pm\!0.26$ & $50.29\!\pm\!0.21$ & $50.06\!\pm\!0.30$ & $49.96\!\pm\!0.18$ & $49.88\!\pm\!0.27$ & $50.33\!\pm\!0.24$ & $49.88\!\pm\!0.14$ \\
RS-Mix & $\boldsymbol{47.49\!\pm\!0.21}$ & $46.91\!\pm\!0.24$ & $46.11\!\pm\!0.20$ & $46.29\!\pm\!0.22$ & $46.13\!\pm\!0.18$ & $45.59\!\pm\!0.24$ & $45.16\!\pm\!0.20$ & $45.22\!\pm\!0.27$ & $44.98\!\pm\!0.24$ & $44.98\!\pm\!0.27$ \\
R-Only & $51.47\!\pm\!0.34$ & $50.26\!\pm\!0.39$ & $49.28\!\pm\!0.45$ & $48.60\!\pm\!0.45$ & $48.38\!\pm\!0.31$ & $47.70\!\pm\!0.28$ & $47.17\!\pm\!0.39$ & $47.20\!\pm\!0.27$ & $47.04\!\pm\!0.23$ & $46.82\!\pm\!0.23$ \\
MC & $56.23\!\pm\!0.09$ & $56.25\!\pm\!0.09$ & $56.30\!\pm\!0.07$ & $56.32\!\pm\!0.05$ & $56.31\!\pm\!0.03$ & $56.32\!\pm\!0.03$ & $56.31\!\pm\!0.02$ & $56.30\!\pm\!0.01$ & $56.26\!\pm\!0.01$ & $56.26\!\pm\!0.00$ \\
CV & $56.48\!\pm\!0.04$ & $56.47\!\pm\!0.03$ & $56.49\!\pm\!0.02$ & $56.47\!\pm\!0.02$ & $56.45\!\pm\!0.02$ & $56.45\!\pm\!0.01$ & $56.44\!\pm\!0.01$ & $56.43\!\pm\!0.01$ & $56.41\!\pm\!0.01$ & $56.41\!\pm\!0.00$ \\
PPI & $56.54\!\pm\!0.04$ & $56.53\!\pm\!0.02$ & $56.55\!\pm\!0.02$ & $56.54\!\pm\!0.01$ & $56.53\!\pm\!0.01$ & $56.52\!\pm\!0.01$ & $56.52\!\pm\!0.01$ & $56.53\!\pm\!0.01$ & $56.53\!\pm\!0.01$ & $56.54\!\pm\!0.00$ \\
\bottomrule
\end{tabular}
}

\confsubtabletitle{TTC$<1$s, average set size}{cov.\ 0.939--0.954}
\resizebox{\textwidth}{!}{%
\begin{tabular}{l|cccccccccc}
\toprule
Method & .1 & .2 & .3 & .4 & .5 & .6 & .7 & .8 & .9 & 1.0 \\
\midrule
\method{} & $\boldsymbol{1.705\!\pm\!0.006}$ & $\boldsymbol{1.664\!\pm\!0.007}$ & $\boldsymbol{1.655\!\pm\!0.004}$ & $\boldsymbol{1.628\!\pm\!0.006}$ & $\boldsymbol{1.623\!\pm\!0.005}$ & $\boldsymbol{1.619\!\pm\!0.004}$ & $\boldsymbol{1.612\!\pm\!0.006}$ & $\boldsymbol{1.608\!\pm\!0.005}$ & $\boldsymbol{1.586\!\pm\!0.006}$ & $\boldsymbol{1.591\!\pm\!0.006}$ \\
PPI-NN & $1.743\!\pm\!0.004$ & $1.727\!\pm\!0.004$ & $1.714\!\pm\!0.002$ & $1.716\!\pm\!0.004$ & $1.711\!\pm\!0.003$ & $1.707\!\pm\!0.004$ & $1.704\!\pm\!0.002$ & $1.702\!\pm\!0.004$ & $1.699\!\pm\!0.003$ & $1.695\!\pm\!0.003$ \\
RS-Mix & $1.770\!\pm\!0.002$ & $1.749\!\pm\!0.004$ & $1.729\!\pm\!0.002$ & $1.715\!\pm\!0.005$ & $1.695\!\pm\!0.004$ & $1.688\!\pm\!0.005$ & $1.686\!\pm\!0.005$ & $1.671\!\pm\!0.003$ & $1.659\!\pm\!0.004$ & $1.660\!\pm\!0.005$ \\
R-Only & $1.736\!\pm\!0.003$ & $1.703\!\pm\!0.004$ & $1.699\!\pm\!0.005$ & $1.686\!\pm\!0.007$ & $1.673\!\pm\!0.005$ & $1.666\!\pm\!0.005$ & $1.663\!\pm\!0.003$ & $1.661\!\pm\!0.004$ & $1.644\!\pm\!0.004$ & $1.643\!\pm\!0.002$ \\
MC & $2.000\!\pm\!0.000$ & $2.000\!\pm\!0.000$ & $2.000\!\pm\!0.000$ & $2.000\!\pm\!0.000$ & $2.000\!\pm\!0.000$ & $2.000\!\pm\!0.000$ & $2.000\!\pm\!0.000$ & $2.000\!\pm\!0.000$ & $2.000\!\pm\!0.000$ & $2.000\!\pm\!0.000$ \\
CV & $2.000\!\pm\!0.000$ & $2.000\!\pm\!0.000$ & $2.000\!\pm\!0.000$ & $2.000\!\pm\!0.000$ & $2.000\!\pm\!0.000$ & $2.000\!\pm\!0.000$ & $2.000\!\pm\!0.000$ & $2.000\!\pm\!0.000$ & $2.000\!\pm\!0.000$ & $2.000\!\pm\!0.000$ \\
PPI & $2.000\!\pm\!0.000$ & $2.000\!\pm\!0.000$ & $2.000\!\pm\!0.000$ & $2.000\!\pm\!0.000$ & $2.000\!\pm\!0.000$ & $2.000\!\pm\!0.000$ & $2.000\!\pm\!0.000$ & $2.000\!\pm\!0.000$ & $2.000\!\pm\!0.000$ & $2.000\!\pm\!0.000$ \\
\bottomrule
\end{tabular}
}

\confsubtabletitle{Drivable Area, average set size}{cov.\ 0.939--0.947}
\resizebox{\textwidth}{!}{%
\begin{tabular}{l|cccccccccc}
\toprule
Method & .1 & .2 & .3 & .4 & .5 & .6 & .7 & .8 & .9 & 1.0 \\
\midrule
\method{} & $1.267\!\pm\!0.009$ & $\boldsymbol{1.227\!\pm\!0.003}$ & $\boldsymbol{1.213\!\pm\!0.005}$ & $\boldsymbol{1.199\!\pm\!0.004}$ & $\boldsymbol{1.191\!\pm\!0.005}$ & $\boldsymbol{1.187\!\pm\!0.003}$ & $\boldsymbol{1.175\!\pm\!0.004}$ & $\boldsymbol{1.179\!\pm\!0.005}$ & $\boldsymbol{1.173\!\pm\!0.003}$ & $\boldsymbol{1.170\!\pm\!0.004}$ \\
PPI-NN & $1.327\!\pm\!0.007$ & $1.307\!\pm\!0.005$ & $1.291\!\pm\!0.006$ & $1.288\!\pm\!0.004$ & $1.285\!\pm\!0.005$ & $1.270\!\pm\!0.004$ & $1.269\!\pm\!0.005$ & $1.271\!\pm\!0.005$ & $1.274\!\pm\!0.005$ & $1.263\!\pm\!0.004$ \\
RS-Mix & $\boldsymbol{1.264\!\pm\!0.004}$ & $1.259\!\pm\!0.002$ & $1.251\!\pm\!0.002$ & $1.247\!\pm\!0.003$ & $1.234\!\pm\!0.004$ & $1.229\!\pm\!0.002$ & $1.225\!\pm\!0.003$ & $1.226\!\pm\!0.004$ & $1.214\!\pm\!0.003$ & $1.208\!\pm\!0.004$ \\
R-Only & $1.325\!\pm\!0.005$ & $1.293\!\pm\!0.005$ & $1.276\!\pm\!0.005$ & $1.260\!\pm\!0.006$ & $1.252\!\pm\!0.003$ & $1.250\!\pm\!0.005$ & $1.242\!\pm\!0.004$ & $1.236\!\pm\!0.003$ & $1.230\!\pm\!0.005$ & $1.220\!\pm\!0.004$ \\
MC & $2.000\!\pm\!0.000$ & $2.000\!\pm\!0.000$ & $2.000\!\pm\!0.000$ & $2.000\!\pm\!0.000$ & $2.000\!\pm\!0.000$ & $2.000\!\pm\!0.000$ & $2.000\!\pm\!0.000$ & $2.000\!\pm\!0.000$ & $2.000\!\pm\!0.000$ & $2.000\!\pm\!0.000$ \\
CV & $2.000\!\pm\!0.000$ & $2.000\!\pm\!0.000$ & $2.000\!\pm\!0.000$ & $2.000\!\pm\!0.000$ & $2.000\!\pm\!0.000$ & $2.000\!\pm\!0.000$ & $2.000\!\pm\!0.000$ & $2.000\!\pm\!0.000$ & $2.000\!\pm\!0.000$ & $2.000\!\pm\!0.000$ \\
PPI & $2.000\!\pm\!0.000$ & $2.000\!\pm\!0.000$ & $2.000\!\pm\!0.000$ & $2.000\!\pm\!0.000$ & $2.000\!\pm\!0.000$ & $2.000\!\pm\!0.000$ & $2.000\!\pm\!0.000$ & $2.000\!\pm\!0.000$ & $2.000\!\pm\!0.000$ & $2.000\!\pm\!0.000$ \\
\bottomrule
\end{tabular}
}

\confsubtabletitle{No At-Fault Collision, average set size}{cov.\ 0.939--0.953}
\resizebox{\textwidth}{!}{%
\begin{tabular}{l|cccccccccc}
\toprule
Method & .1 & .2 & .3 & .4 & .5 & .6 & .7 & .8 & .9 & 1.0 \\
\midrule
\method{} & $\boldsymbol{1.693\!\pm\!0.004}$ & $\boldsymbol{1.649\!\pm\!0.007}$ & $\boldsymbol{1.637\!\pm\!0.005}$ & $\boldsymbol{1.619\!\pm\!0.007}$ & $\boldsymbol{1.609\!\pm\!0.007}$ & $\boldsymbol{1.601\!\pm\!0.004}$ & $\boldsymbol{1.600\!\pm\!0.004}$ & $\boldsymbol{1.586\!\pm\!0.005}$ & $\boldsymbol{1.570\!\pm\!0.004}$ & $\boldsymbol{1.571\!\pm\!0.005}$ \\
PPI-NN & $1.724\!\pm\!0.004$ & $1.714\!\pm\!0.004$ & $1.708\!\pm\!0.003$ & $1.706\!\pm\!0.005$ & $1.695\!\pm\!0.004$ & $1.695\!\pm\!0.004$ & $1.701\!\pm\!0.002$ & $1.694\!\pm\!0.004$ & $1.690\!\pm\!0.004$ & $1.687\!\pm\!0.005$ \\
RS-Mix & $1.768\!\pm\!0.003$ & $1.747\!\pm\!0.004$ & $1.726\!\pm\!0.004$ & $1.722\!\pm\!0.003$ & $1.701\!\pm\!0.007$ & $1.689\!\pm\!0.007$ & $1.669\!\pm\!0.005$ & $1.664\!\pm\!0.005$ & $1.642\!\pm\!0.004$ & $1.638\!\pm\!0.004$ \\
R-Only & $1.720\!\pm\!0.005$ & $1.686\!\pm\!0.005$ & $1.686\!\pm\!0.006$ & $1.669\!\pm\!0.003$ & $1.662\!\pm\!0.004$ & $1.652\!\pm\!0.003$ & $1.649\!\pm\!0.003$ & $1.646\!\pm\!0.004$ & $1.637\!\pm\!0.004$ & $1.631\!\pm\!0.005$ \\
MC & $2.000\!\pm\!0.000$ & $2.000\!\pm\!0.000$ & $2.000\!\pm\!0.000$ & $2.000\!\pm\!0.000$ & $2.000\!\pm\!0.000$ & $2.000\!\pm\!0.000$ & $2.000\!\pm\!0.000$ & $2.000\!\pm\!0.000$ & $2.000\!\pm\!0.000$ & $2.000\!\pm\!0.000$ \\
CV & $2.000\!\pm\!0.000$ & $2.000\!\pm\!0.000$ & $2.000\!\pm\!0.000$ & $2.000\!\pm\!0.000$ & $2.000\!\pm\!0.000$ & $2.000\!\pm\!0.000$ & $2.000\!\pm\!0.000$ & $2.000\!\pm\!0.000$ & $2.000\!\pm\!0.000$ & $2.000\!\pm\!0.000$ \\
PPI & $2.000\!\pm\!0.000$ & $2.000\!\pm\!0.000$ & $2.000\!\pm\!0.000$ & $2.000\!\pm\!0.000$ & $2.000\!\pm\!0.000$ & $2.000\!\pm\!0.000$ & $2.000\!\pm\!0.000$ & $2.000\!\pm\!0.000$ & $2.000\!\pm\!0.000$ & $2.000\!\pm\!0.000$ \\
\bottomrule
\end{tabular}
}
\vspace{-6pt}
\end{table}

\begin{table}[H]
\centering
\vspace{-4pt}
\confblockcaption
{Full split conformal width sweep on nuPlan Simple Vector.}
{For ADE, width is interval length in meters; for binary targets, width is average prediction-set size.}
{tab:conformal-simple-vector-full}
{ADE, interval width (m)}
{cov.\ 0.949--0.951}
\resizebox{\textwidth}{!}{%
\begin{tabular}{l|cccccccccc}
\toprule
Method & .1 & .2 & .3 & .4 & .5 & .6 & .7 & .8 & .9 & 1.0 \\
\midrule
\method{} & $67.63\!\pm\!0.51$ & $\boldsymbol{66.47\!\pm\!0.33}$ & $66.63\!\pm\!0.23$ & $\boldsymbol{66.21\!\pm\!0.20}$ & $\boldsymbol{65.63\!\pm\!0.18}$ & $\boldsymbol{65.94\!\pm\!0.33}$ & $\boldsymbol{65.67\!\pm\!0.39}$ & $\boldsymbol{65.73\!\pm\!0.27}$ & $\boldsymbol{65.26\!\pm\!0.27}$ & $\boldsymbol{65.20\!\pm\!0.26}$ \\
PPI-NN & $70.42\!\pm\!0.24$ & $68.72\!\pm\!0.35$ & $68.72\!\pm\!0.39$ & $67.32\!\pm\!0.45$ & $67.94\!\pm\!0.32$ & $68.18\!\pm\!0.22$ & $67.80\!\pm\!0.37$ & $67.99\!\pm\!0.30$ & $67.54\!\pm\!0.37$ & $67.23\!\pm\!0.23$ \\
RS-Mix & $\boldsymbol{66.75\!\pm\!0.24}$ & $67.13\!\pm\!0.27$ & $\boldsymbol{66.35\!\pm\!0.26}$ & $66.59\!\pm\!0.28$ & $66.80\!\pm\!0.39$ & $66.17\!\pm\!0.24$ & $66.36\!\pm\!0.45$ & $66.38\!\pm\!0.18$ & $66.78\!\pm\!0.36$ & $65.82\!\pm\!0.34$ \\
R-Only & $69.54\!\pm\!0.45$ & $68.61\!\pm\!0.55$ & $67.53\!\pm\!0.35$ & $67.64\!\pm\!0.36$ & $67.53\!\pm\!0.37$ & $66.82\!\pm\!0.34$ & $66.90\!\pm\!0.29$ & $66.57\!\pm\!0.16$ & $66.62\!\pm\!0.18$ & $65.90\!\pm\!0.21$ \\
MC & $105.42\!\pm\!0.16$ & $105.26\!\pm\!0.10$ & $105.25\!\pm\!0.08$ & $105.24\!\pm\!0.07$ & $105.28\!\pm\!0.06$ & $105.27\!\pm\!0.04$ & $105.27\!\pm\!0.03$ & $105.22\!\pm\!0.02$ & $105.19\!\pm\!0.01$ & $105.18\!\pm\!0.00$ \\
CV & $105.49\!\pm\!0.09$ & $105.36\!\pm\!0.05$ & $105.31\!\pm\!0.04$ & $105.29\!\pm\!0.04$ & $105.31\!\pm\!0.03$ & $105.31\!\pm\!0.03$ & $105.33\!\pm\!0.02$ & $105.29\!\pm\!0.01$ & $105.28\!\pm\!0.01$ & $105.28\!\pm\!0.00$ \\
PPI & $105.50\!\pm\!0.10$ & $105.39\!\pm\!0.06$ & $105.33\!\pm\!0.05$ & $105.31\!\pm\!0.05$ & $105.32\!\pm\!0.03$ & $105.34\!\pm\!0.03$ & $105.37\!\pm\!0.02$ & $105.35\!\pm\!0.01$ & $105.35\!\pm\!0.01$ & $105.36\!\pm\!0.00$ \\
\bottomrule
\end{tabular}
}

\confsubtabletitle{TTC$<1$s, average set size}{cov.\ 0.942--0.953}
\resizebox{\textwidth}{!}{%
\begin{tabular}{l|cccccccccc}
\toprule
Method & .1 & .2 & .3 & .4 & .5 & .6 & .7 & .8 & .9 & 1.0 \\
\midrule
\method{} & $\boldsymbol{1.721\!\pm\!0.005}$ & $\boldsymbol{1.697\!\pm\!0.005}$ & $\boldsymbol{1.682\!\pm\!0.008}$ & $\boldsymbol{1.688\!\pm\!0.006}$ & $\boldsymbol{1.675\!\pm\!0.004}$ & $\boldsymbol{1.662\!\pm\!0.005}$ & $\boldsymbol{1.659\!\pm\!0.003}$ & $\boldsymbol{1.663\!\pm\!0.003}$ & $\boldsymbol{1.651\!\pm\!0.004}$ & $\boldsymbol{1.651\!\pm\!0.003}$ \\
PPI-NN & $1.745\!\pm\!0.004$ & $1.715\!\pm\!0.005$ & $1.707\!\pm\!0.003$ & $1.703\!\pm\!0.004$ & $1.706\!\pm\!0.004$ & $1.698\!\pm\!0.003$ & $1.686\!\pm\!0.003$ & $1.688\!\pm\!0.004$ & $1.692\!\pm\!0.003$ & $1.683\!\pm\!0.004$ \\
RS-Mix & $1.723\!\pm\!0.002$ & $1.712\!\pm\!0.003$ & $1.707\!\pm\!0.003$ & $1.694\!\pm\!0.003$ & $1.688\!\pm\!0.002$ & $1.683\!\pm\!0.004$ & $1.670\!\pm\!0.005$ & $1.675\!\pm\!0.005$ & $1.674\!\pm\!0.002$ & $1.662\!\pm\!0.002$ \\
R-Only & $1.738\!\pm\!0.007$ & $1.720\!\pm\!0.004$ & $1.718\!\pm\!0.007$ & $1.709\!\pm\!0.006$ & $1.701\!\pm\!0.004$ & $1.694\!\pm\!0.003$ & $1.688\!\pm\!0.004$ & $1.689\!\pm\!0.004$ & $1.674\!\pm\!0.003$ & $1.677\!\pm\!0.004$ \\
MC & $2.000\!\pm\!0.000$ & $2.000\!\pm\!0.000$ & $2.000\!\pm\!0.000$ & $2.000\!\pm\!0.000$ & $2.000\!\pm\!0.000$ & $2.000\!\pm\!0.000$ & $2.000\!\pm\!0.000$ & $2.000\!\pm\!0.000$ & $2.000\!\pm\!0.000$ & $2.000\!\pm\!0.000$ \\
CV & $2.000\!\pm\!0.000$ & $2.000\!\pm\!0.000$ & $2.000\!\pm\!0.000$ & $2.000\!\pm\!0.000$ & $2.000\!\pm\!0.000$ & $2.000\!\pm\!0.000$ & $2.000\!\pm\!0.000$ & $2.000\!\pm\!0.000$ & $2.000\!\pm\!0.000$ & $2.000\!\pm\!0.000$ \\
PPI & $2.000\!\pm\!0.000$ & $2.000\!\pm\!0.000$ & $2.000\!\pm\!0.000$ & $2.000\!\pm\!0.000$ & $2.000\!\pm\!0.000$ & $2.000\!\pm\!0.000$ & $2.000\!\pm\!0.000$ & $2.000\!\pm\!0.000$ & $2.000\!\pm\!0.000$ & $2.000\!\pm\!0.000$ \\
\bottomrule
\end{tabular}
}

\confsubtabletitle{Drivable Area, average set size}{cov.\ 0.953--0.956}
\resizebox{\textwidth}{!}{%
\begin{tabular}{l|cccccccccc}
\toprule
Method & .1 & .2 & .3 & .4 & .5 & .6 & .7 & .8 & .9 & 1.0 \\
\midrule
\method{} & $1.350\!\pm\!0.004$ & $\boldsymbol{1.328\!\pm\!0.004}$ & $\boldsymbol{1.320\!\pm\!0.003}$ & $\boldsymbol{1.306\!\pm\!0.004}$ & $\boldsymbol{1.306\!\pm\!0.004}$ & $\boldsymbol{1.295\!\pm\!0.004}$ & $\boldsymbol{1.291\!\pm\!0.003}$ & $\boldsymbol{1.284\!\pm\!0.004}$ & $\boldsymbol{1.284\!\pm\!0.003}$ & $\boldsymbol{1.281\!\pm\!0.005}$ \\
PPI-NN & $1.390\!\pm\!0.004$ & $1.377\!\pm\!0.004$ & $1.369\!\pm\!0.004$ & $1.356\!\pm\!0.003$ & $1.354\!\pm\!0.004$ & $1.349\!\pm\!0.003$ & $1.354\!\pm\!0.003$ & $1.346\!\pm\!0.004$ & $1.351\!\pm\!0.004$ & $1.346\!\pm\!0.004$ \\
RS-Mix & $\boldsymbol{1.337\!\pm\!0.005}$ & $1.333\!\pm\!0.002$ & $1.327\!\pm\!0.004$ & $1.317\!\pm\!0.004$ & $1.309\!\pm\!0.004$ & $1.298\!\pm\!0.004$ & $1.300\!\pm\!0.004$ & $1.296\!\pm\!0.004$ & $1.294\!\pm\!0.004$ & $1.288\!\pm\!0.002$ \\
R-Only & $1.375\!\pm\!0.004$ & $1.355\!\pm\!0.004$ & $1.341\!\pm\!0.004$ & $1.331\!\pm\!0.005$ & $1.331\!\pm\!0.006$ & $1.315\!\pm\!0.004$ & $1.317\!\pm\!0.003$ & $1.308\!\pm\!0.003$ & $1.305\!\pm\!0.004$ & $1.304\!\pm\!0.004$ \\
MC & $2.000\!\pm\!0.000$ & $2.000\!\pm\!0.000$ & $2.000\!\pm\!0.000$ & $2.000\!\pm\!0.000$ & $2.000\!\pm\!0.000$ & $2.000\!\pm\!0.000$ & $2.000\!\pm\!0.000$ & $2.000\!\pm\!0.000$ & $2.000\!\pm\!0.000$ & $2.000\!\pm\!0.000$ \\
CV & $2.000\!\pm\!0.000$ & $2.000\!\pm\!0.000$ & $2.000\!\pm\!0.000$ & $2.000\!\pm\!0.000$ & $2.000\!\pm\!0.000$ & $2.000\!\pm\!0.000$ & $2.000\!\pm\!0.000$ & $2.000\!\pm\!0.000$ & $2.000\!\pm\!0.000$ & $2.000\!\pm\!0.000$ \\
PPI & $2.000\!\pm\!0.000$ & $2.000\!\pm\!0.000$ & $2.000\!\pm\!0.000$ & $2.000\!\pm\!0.000$ & $2.000\!\pm\!0.000$ & $2.000\!\pm\!0.000$ & $2.000\!\pm\!0.000$ & $2.000\!\pm\!0.000$ & $2.000\!\pm\!0.000$ & $2.000\!\pm\!0.000$ \\
\bottomrule
\end{tabular}
}

\confsubtabletitle{No At-Fault Collision, average set size}{cov.\ 0.939--0.951}
\resizebox{\textwidth}{!}{%
\begin{tabular}{l|cccccccccc}
\toprule
Method & .1 & .2 & .3 & .4 & .5 & .6 & .7 & .8 & .9 & 1.0 \\
\midrule
\method{} & $1.716\!\pm\!0.006$ & $\boldsymbol{1.691\!\pm\!0.004}$ & $\boldsymbol{1.684\!\pm\!0.006}$ & $1.676\!\pm\!0.007$ & $1.667\!\pm\!0.004$ & $1.665\!\pm\!0.006$ & $\boldsymbol{1.645\!\pm\!0.003}$ & $\boldsymbol{1.640\!\pm\!0.005}$ & $\boldsymbol{1.638\!\pm\!0.003}$ & $\boldsymbol{1.633\!\pm\!0.003}$ \\
PPI-NN & $1.732\!\pm\!0.004$ & $1.714\!\pm\!0.005$ & $1.704\!\pm\!0.005$ & $1.694\!\pm\!0.006$ & $1.687\!\pm\!0.002$ & $1.687\!\pm\!0.003$ & $1.676\!\pm\!0.004$ & $1.684\!\pm\!0.004$ & $1.678\!\pm\!0.004$ & $1.670\!\pm\!0.002$ \\
RS-Mix & $\boldsymbol{1.710\!\pm\!0.004}$ & $1.697\!\pm\!0.003$ & $1.689\!\pm\!0.003$ & $\boldsymbol{1.673\!\pm\!0.003}$ & $\boldsymbol{1.664\!\pm\!0.004}$ & $\boldsymbol{1.659\!\pm\!0.003}$ & $1.647\!\pm\!0.004$ & $1.642\!\pm\!0.004$ & $1.647\!\pm\!0.003$ & $1.634\!\pm\!0.003$ \\
R-Only & $1.737\!\pm\!0.006$ & $1.722\!\pm\!0.004$ & $1.715\!\pm\!0.006$ & $1.713\!\pm\!0.006$ & $1.704\!\pm\!0.005$ & $1.694\!\pm\!0.006$ & $1.692\!\pm\!0.004$ & $1.690\!\pm\!0.005$ & $1.672\!\pm\!0.005$ & $1.679\!\pm\!0.004$ \\
MC & $2.000\!\pm\!0.000$ & $2.000\!\pm\!0.000$ & $2.000\!\pm\!0.000$ & $2.000\!\pm\!0.000$ & $2.000\!\pm\!0.000$ & $2.000\!\pm\!0.000$ & $2.000\!\pm\!0.000$ & $2.000\!\pm\!0.000$ & $2.000\!\pm\!0.000$ & $2.000\!\pm\!0.000$ \\
CV & $2.000\!\pm\!0.000$ & $2.000\!\pm\!0.000$ & $2.000\!\pm\!0.000$ & $2.000\!\pm\!0.000$ & $2.000\!\pm\!0.000$ & $2.000\!\pm\!0.000$ & $2.000\!\pm\!0.000$ & $2.000\!\pm\!0.000$ & $2.000\!\pm\!0.000$ & $2.000\!\pm\!0.000$ \\
PPI & $2.000\!\pm\!0.000$ & $2.000\!\pm\!0.000$ & $2.000\!\pm\!0.000$ & $2.000\!\pm\!0.000$ & $2.000\!\pm\!0.000$ & $2.000\!\pm\!0.000$ & $2.000\!\pm\!0.000$ & $2.000\!\pm\!0.000$ & $2.000\!\pm\!0.000$ & $2.000\!\pm\!0.000$ \\
\bottomrule
\end{tabular}
}
\vspace{-6pt}
\end{table}

\begin{table}[H]
\centering
\vspace{-4pt}
\confblockcaption
{Full split conformal width sweep on Go2 Sim2Sim.}
{Width is regression interval length reported as \(\times 10^{-2}\).}
{tab:conformal-go2-full}
{Velocity, interval width \(\times 10^{-2}\)}
{cov.\ 0.945--0.955}
\resizebox{\textwidth}{!}{%
\begin{tabular}{l|cccccccccc}
\toprule
Method & .1 & .2 & .3 & .4 & .5 & .6 & .7 & .8 & .9 & 1.0 \\
\midrule
\method{} & $\boldsymbol{11.37\!\pm\!0.34}$ & $\boldsymbol{11.05\!\pm\!0.48}$ & $\boldsymbol{11.05\!\pm\!0.41}$ & $\boldsymbol{11.04\!\pm\!0.48}$ & $\boldsymbol{11.08\!\pm\!0.59}$ & $\boldsymbol{10.98\!\pm\!0.46}$ & $\boldsymbol{10.97\!\pm\!0.50}$ & $\boldsymbol{11.02\!\pm\!0.58}$ & $\boldsymbol{10.75\!\pm\!0.55}$ & $\boldsymbol{10.84\!\pm\!0.48}$ \\
PPI-NN & $12.42\!\pm\!0.83$ & $12.57\!\pm\!1.27$ & $11.16\!\pm\!0.44$ & $11.75\!\pm\!1.00$ & $11.26\!\pm\!0.45$ & $12.01\!\pm\!0.51$ & $11.39\!\pm\!0.57$ & $11.55\!\pm\!0.39$ & $11.14\!\pm\!0.34$ & $11.42\!\pm\!0.38$ \\
RS-Mix & $12.39\!\pm\!0.36$ & $12.61\!\pm\!0.44$ & $12.39\!\pm\!0.40$ & $12.10\!\pm\!0.33$ & $12.34\!\pm\!0.39$ & $12.34\!\pm\!0.34$ & $12.29\!\pm\!0.45$ & $12.07\!\pm\!0.29$ & $12.16\!\pm\!0.43$ & $11.56\!\pm\!0.28$ \\
R-Only & $12.05\!\pm\!0.42$ & $12.11\!\pm\!0.47$ & $11.44\!\pm\!0.47$ & $11.57\!\pm\!0.49$ & $11.53\!\pm\!0.47$ & $11.27\!\pm\!0.40$ & $11.06\!\pm\!0.52$ & $11.45\!\pm\!0.47$ & $10.86\!\pm\!0.48$ & $11.12\!\pm\!0.59$ \\
MC & $13.18\!\pm\!0.84$ & $13.02\!\pm\!0.92$ & $12.51\!\pm\!0.88$ & $12.85\!\pm\!0.87$ & $12.84\!\pm\!0.90$ & $12.67\!\pm\!0.83$ & $12.84\!\pm\!0.86$ & $12.60\!\pm\!0.87$ & $12.64\!\pm\!0.88$ & $12.64\!\pm\!0.88$ \\
CV & $12.82\!\pm\!0.85$ & $12.57\!\pm\!0.87$ & $12.46\!\pm\!0.87$ & $12.57\!\pm\!0.85$ & $12.51\!\pm\!0.85$ & $12.55\!\pm\!0.81$ & $12.57\!\pm\!0.87$ & $12.48\!\pm\!0.85$ & $12.51\!\pm\!0.87$ & $12.50\!\pm\!0.88$ \\
PPI & $12.53\!\pm\!0.85$ & $12.18\!\pm\!0.72$ & $12.51\!\pm\!0.82$ & $12.27\!\pm\!0.81$ & $12.26\!\pm\!0.76$ & $12.40\!\pm\!0.79$ & $12.41\!\pm\!0.82$ & $12.39\!\pm\!0.82$ & $12.44\!\pm\!0.84$ & $12.41\!\pm\!0.84$ \\
\bottomrule
\end{tabular}
}

\confsubtabletitle{Yaw Rate, interval width \(\times 10^{-2}\)}{cov.\ 0.940--0.943}
\resizebox{\textwidth}{!}{%
\begin{tabular}{l|cccccccccc}
\toprule
Method & .1 & .2 & .3 & .4 & .5 & .6 & .7 & .8 & .9 & 1.0 \\
\midrule
\method{} & $11.31\!\pm\!0.51$ & $\boldsymbol{9.77\!\pm\!0.30}$ & $\boldsymbol{9.00\!\pm\!0.27}$ & $\boldsymbol{8.62\!\pm\!0.30}$ & $\boldsymbol{8.24\!\pm\!0.36}$ & $\boldsymbol{8.12\!\pm\!0.33}$ & $\boldsymbol{8.05\!\pm\!0.39}$ & $\boldsymbol{7.95\!\pm\!0.35}$ & $\boldsymbol{7.69\!\pm\!0.37}$ & $\boldsymbol{7.60\!\pm\!0.39}$ \\
PPI-NN & $12.04\!\pm\!0.31$ & $12.19\!\pm\!0.13$ & $11.67\!\pm\!0.46$ & $10.47\!\pm\!0.40$ & $11.04\!\pm\!0.30$ & $10.58\!\pm\!0.24$ & $10.75\!\pm\!0.30$ & $10.44\!\pm\!0.16$ & $10.32\!\pm\!0.21$ & $10.27\!\pm\!0.31$ \\
RS-Mix & $\boldsymbol{10.60\!\pm\!0.28}$ & $10.62\!\pm\!0.24$ & $10.48\!\pm\!0.21$ & $10.36\!\pm\!0.26$ & $10.07\!\pm\!0.31$ & $9.76\!\pm\!0.24$ & $9.77\!\pm\!0.19$ & $9.63\!\pm\!0.27$ & $9.80\!\pm\!0.28$ & $9.12\!\pm\!0.22$ \\
R-Only & $12.21\!\pm\!0.32$ & $12.42\!\pm\!0.36$ & $11.35\!\pm\!0.29$ & $10.18\!\pm\!0.26$ & $9.29\!\pm\!0.28$ & $8.89\!\pm\!0.34$ & $8.52\!\pm\!0.23$ & $8.19\!\pm\!0.38$ & $7.99\!\pm\!0.34$ & $7.75\!\pm\!0.31$ \\
MC & $12.11\!\pm\!0.36$ & $12.04\!\pm\!0.45$ & $11.86\!\pm\!0.33$ & $12.04\!\pm\!0.33$ & $12.03\!\pm\!0.25$ & $12.13\!\pm\!0.30$ & $11.94\!\pm\!0.30$ & $11.88\!\pm\!0.30$ & $11.95\!\pm\!0.31$ & $11.95\!\pm\!0.31$ \\
CV & $11.96\!\pm\!0.39$ & $11.85\!\pm\!0.40$ & $11.86\!\pm\!0.34$ & $11.80\!\pm\!0.27$ & $11.83\!\pm\!0.24$ & $12.03\!\pm\!0.29$ & $11.84\!\pm\!0.28$ & $11.82\!\pm\!0.28$ & $11.86\!\pm\!0.30$ & $11.88\!\pm\!0.29$ \\
PPI & $12.14\!\pm\!0.48$ & $11.94\!\pm\!0.31$ & $11.86\!\pm\!0.34$ & $11.62\!\pm\!0.20$ & $11.78\!\pm\!0.24$ & $11.99\!\pm\!0.30$ & $11.72\!\pm\!0.27$ & $11.75\!\pm\!0.28$ & $11.76\!\pm\!0.27$ & $11.75\!\pm\!0.29$ \\
\bottomrule
\end{tabular}
}
\vspace{-6pt}
\end{table}

\subsection{Additional OOD Generalization Results}
\label{app:ood results}

\begin{figure}[H]
    \centering
    \vspace{-8pt}
    \captionsetup{skip=3pt}

    \begin{subfigure}[t]{0.48\linewidth}
        \centering
        \includegraphics[width=\linewidth]{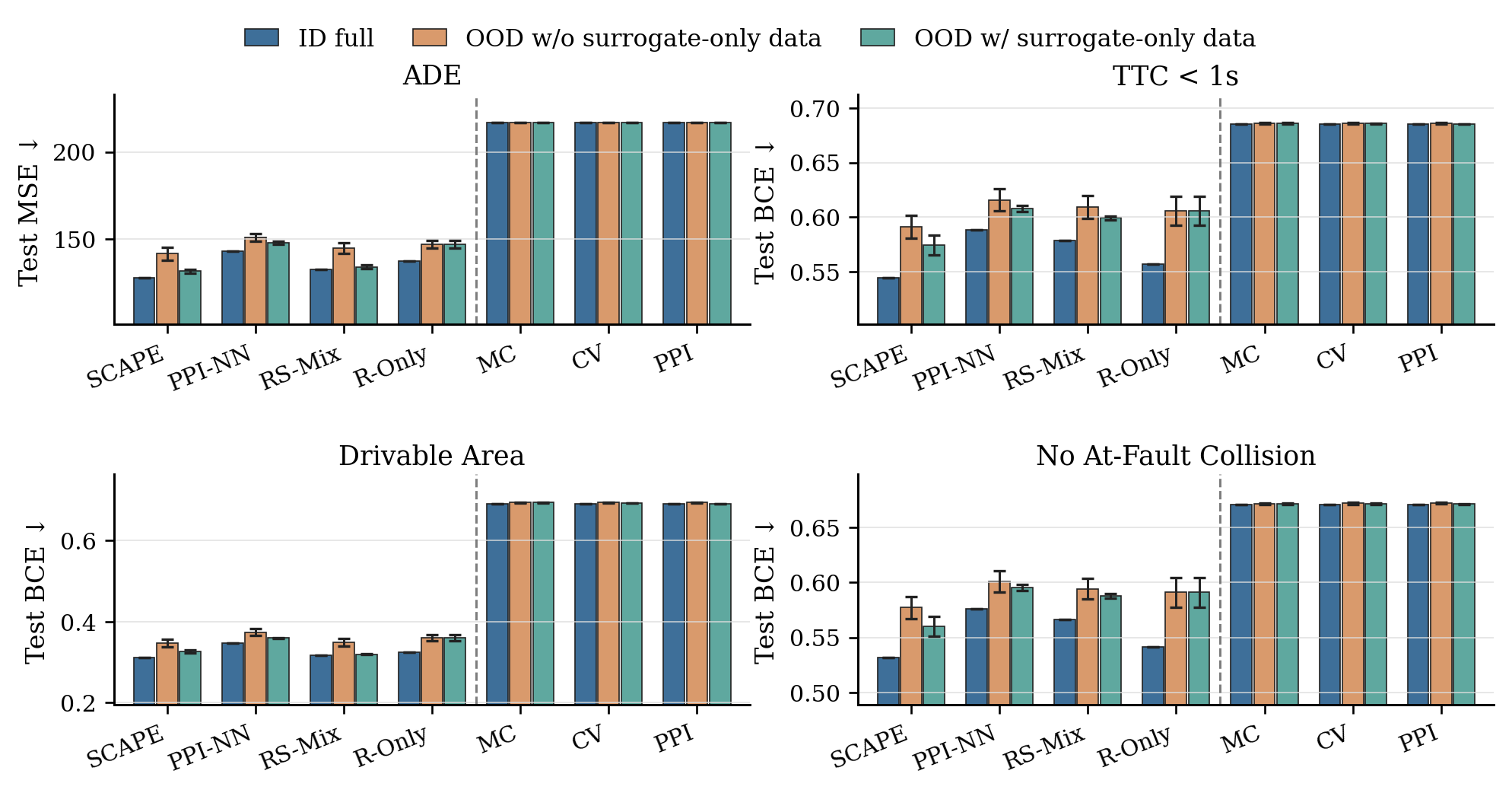}
        \caption{Vector}
        \label{fig:app-ood-vector}
    \end{subfigure}
    \hfill
    \begin{subfigure}[t]{0.48\linewidth}
        \centering
        \includegraphics[width=\linewidth]{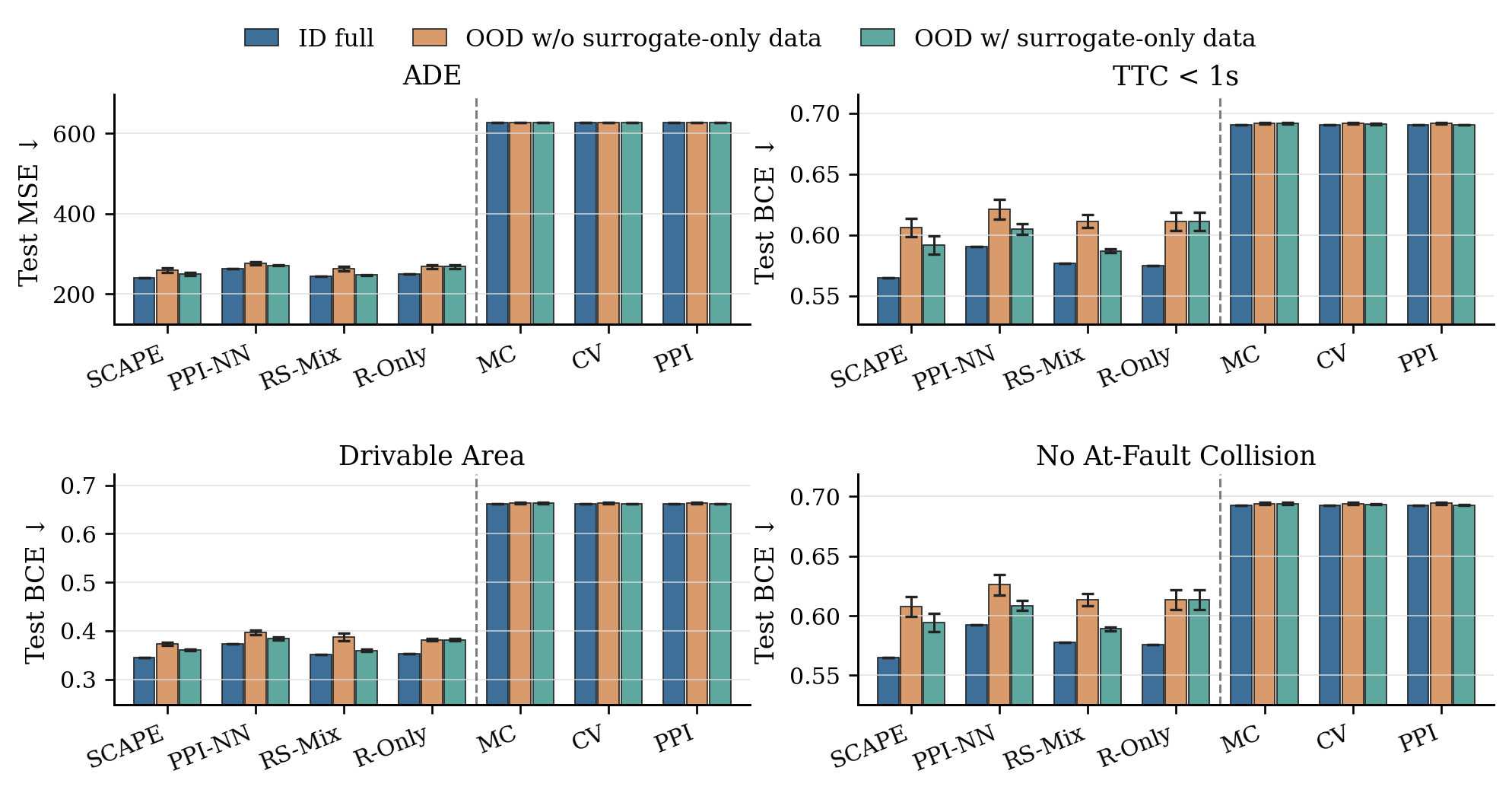}
        \caption{Simple Vector}
        \label{fig:app-ood-simple}
    \end{subfigure}

    \caption{\small \textbf{Additional city-level OOD generalization results on nuPlan.}
    Panels (a) and (b) report results for the Vector and Simple Vector planners, respectively.
    Results compare full training, OOD training with surrogate-only data from the removed city, and OOD training without removed-city surrogate data.
    Results are averaged over four leave-one-city-out settings.}
    \label{fig:app-ood-additional}
    \vspace{-8pt}
\end{figure}

\subsection{Full Surrogate Correction Results}
\label{app:correction results}

\begin{figure}[H]
    \centering
    \vspace{-8pt}
    \captionsetup{skip=3pt}

    \begin{subfigure}[t]{0.32\linewidth}
        \centering
        \includegraphics[width=\linewidth]{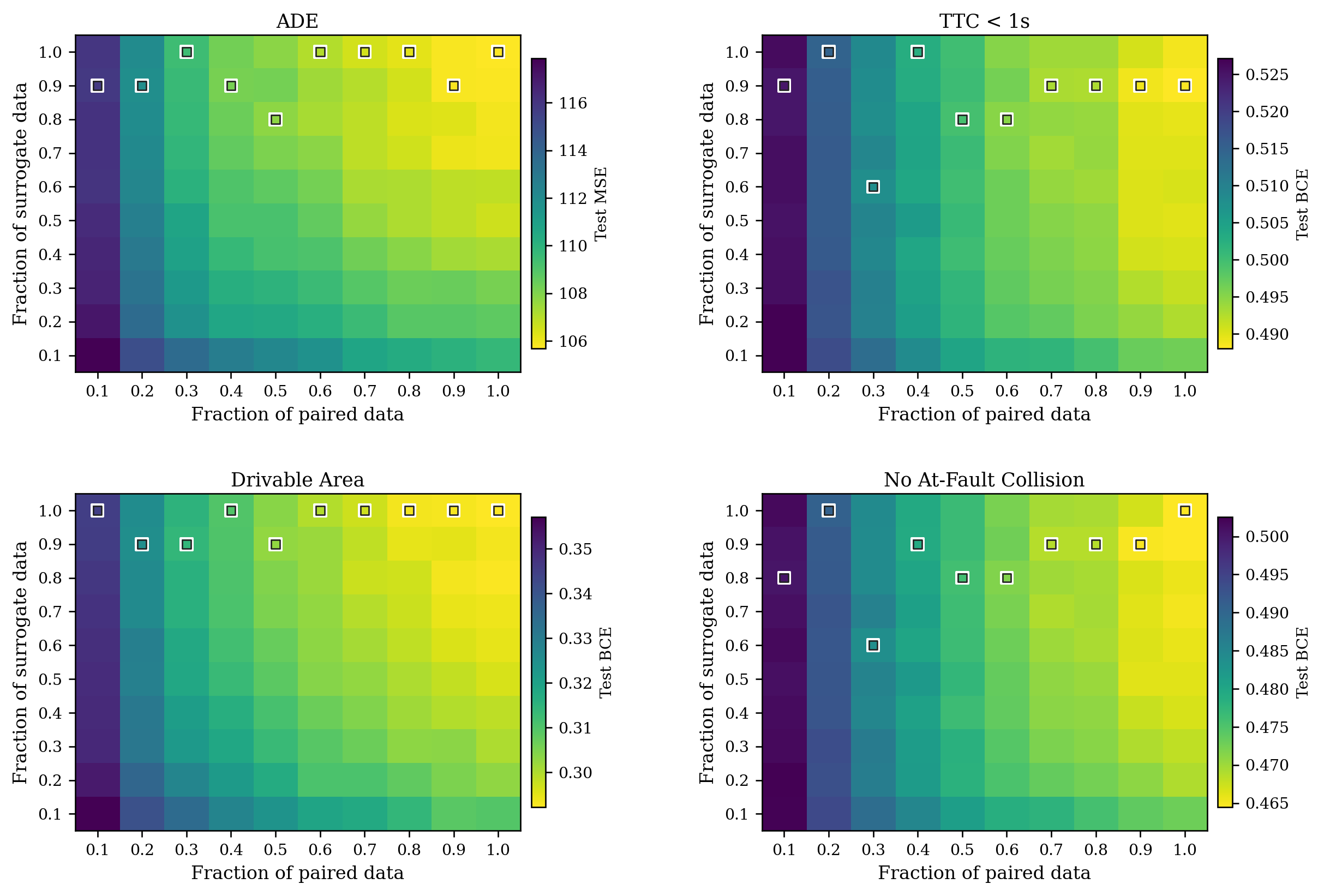}
        \caption{\method{}}
        \label{fig:app-correction-scape}
    \end{subfigure}
    \hfill
    \begin{subfigure}[t]{0.32\linewidth}
        \centering
        \includegraphics[width=\linewidth]{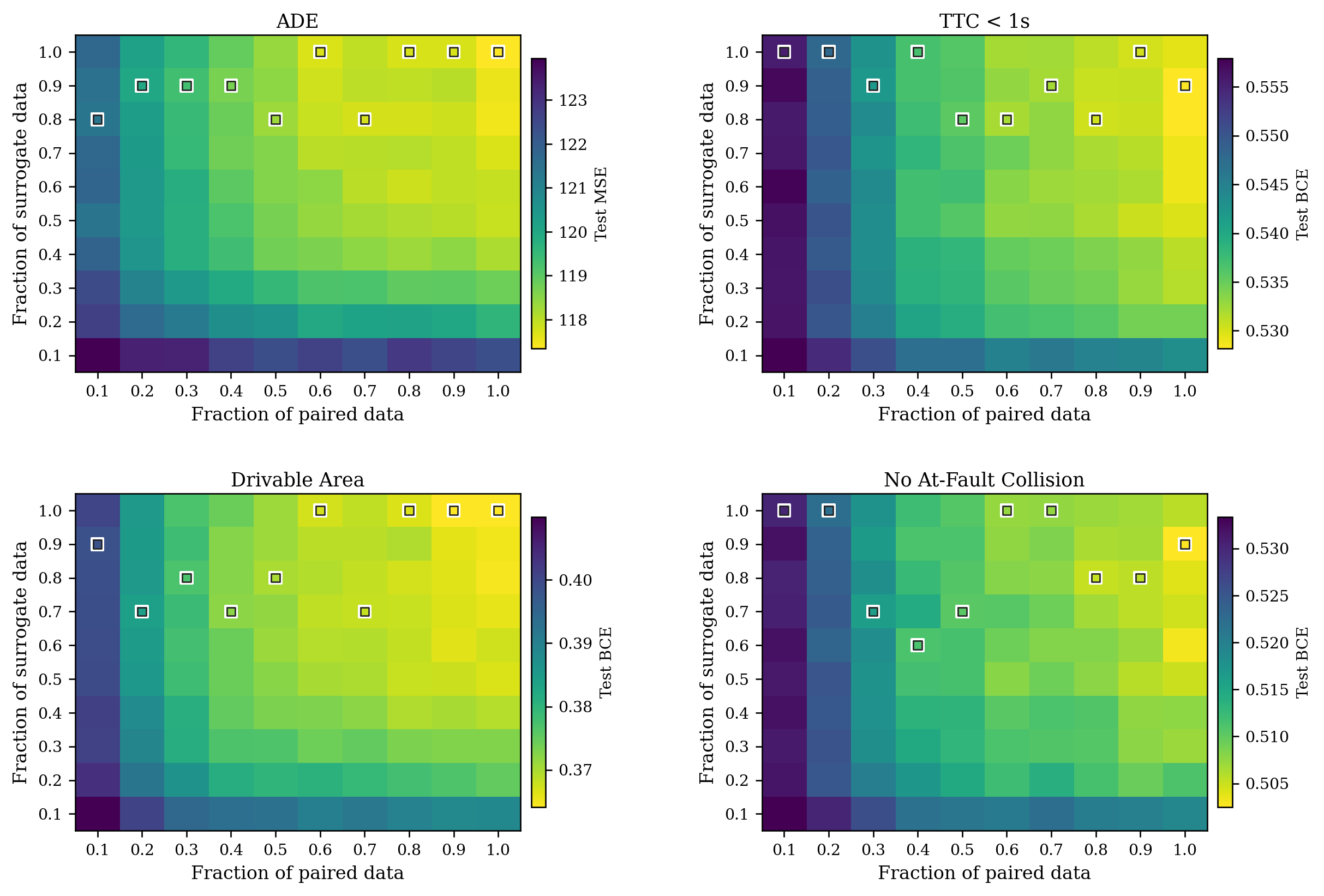}
        \caption{PPI-NN}
        \label{fig:app-correction-ppi-nn}
    \end{subfigure}
    \hfill
    \begin{subfigure}[t]{0.32\linewidth}
        \centering
        \includegraphics[width=\linewidth]{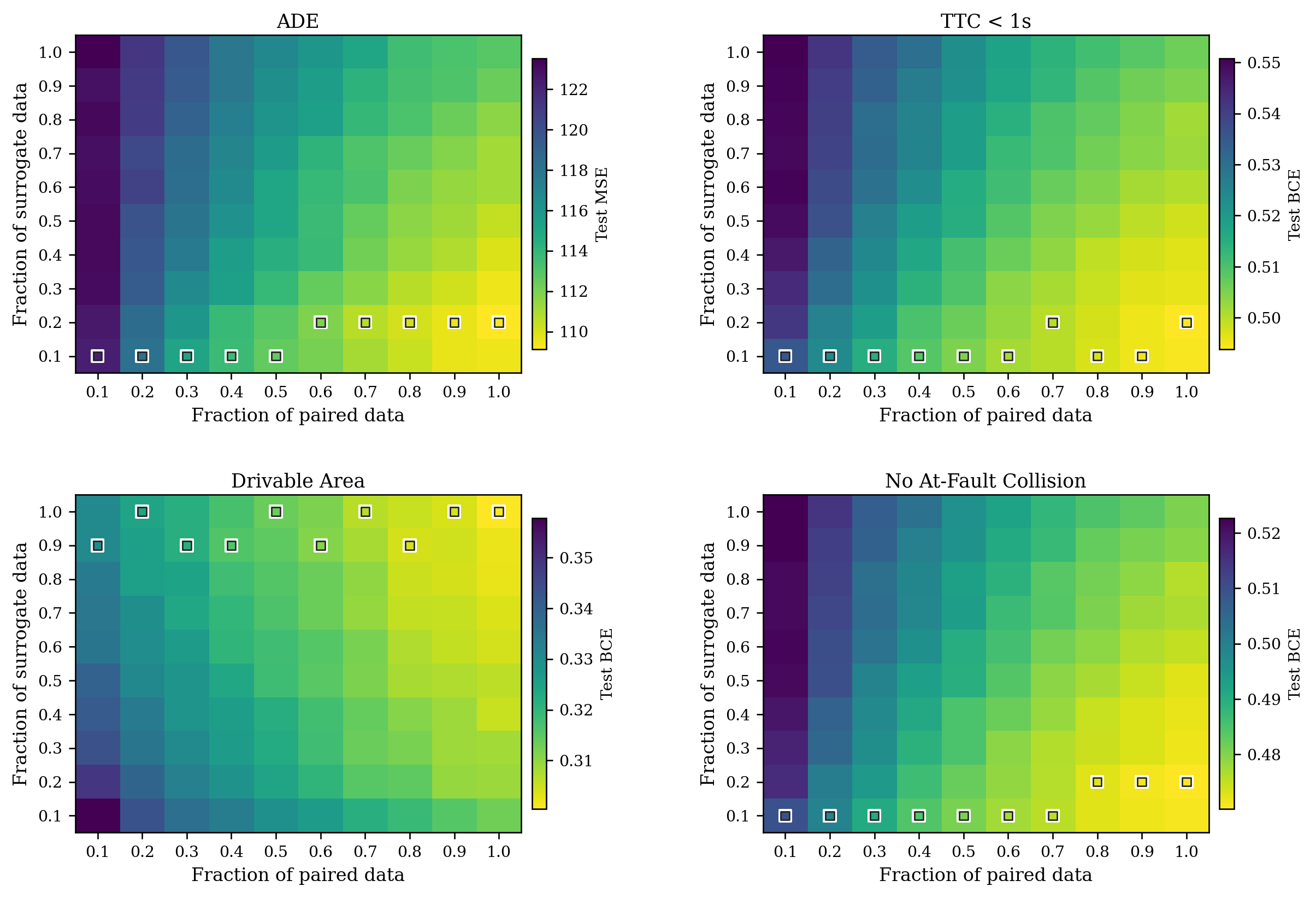}
        \caption{RS-Mix}
        \label{fig:app-correction-rs-mix}
    \end{subfigure}

    \caption{\small \textbf{Surrogate scaling and correction capacity on nuPlan Urban Driver.}
    Paired-data and surrogate-only fractions are varied jointly for \method{}, PPI-NN, and RS-Mix.
    Results are averaged over \(3\times3\) paired-data and surrogate-data seeds.
    Square markers denote the best surrogate-only fraction for each paired-data fraction.}
    \label{fig:app-correction-all}
    \vspace{-8pt}
\end{figure}

\subsection{Additional Evaluator-Guided Planner Selection Results}
\label{app:decision making appendix}

\begin{figure}[H]
    \centering
    \vspace{-8pt}
    \captionsetup{skip=3pt}

    \begin{subfigure}[t]{0.48\linewidth}
        \centering
        \includegraphics[width=\linewidth]{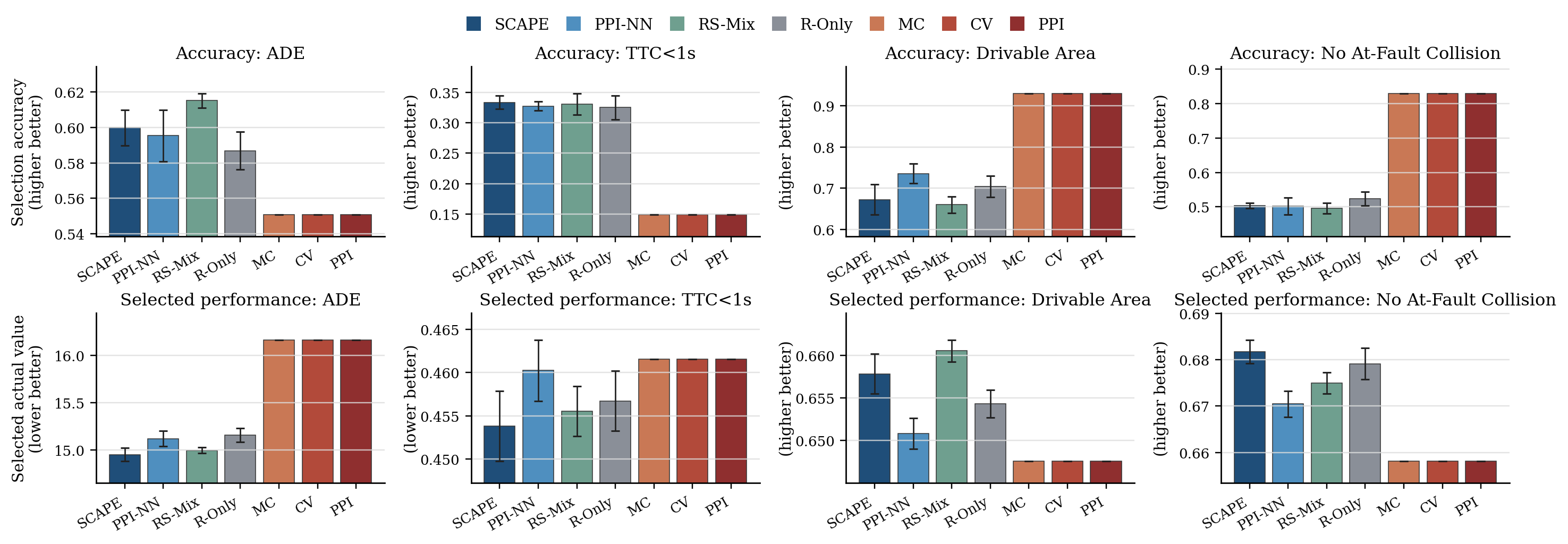}
        \caption{\(\rho_p=0.1\)}
        \label{fig:policy_moe_frac_01}
    \end{subfigure}
    \hfill
    \begin{subfigure}[t]{0.48\linewidth}
        \centering
        \includegraphics[width=\linewidth]{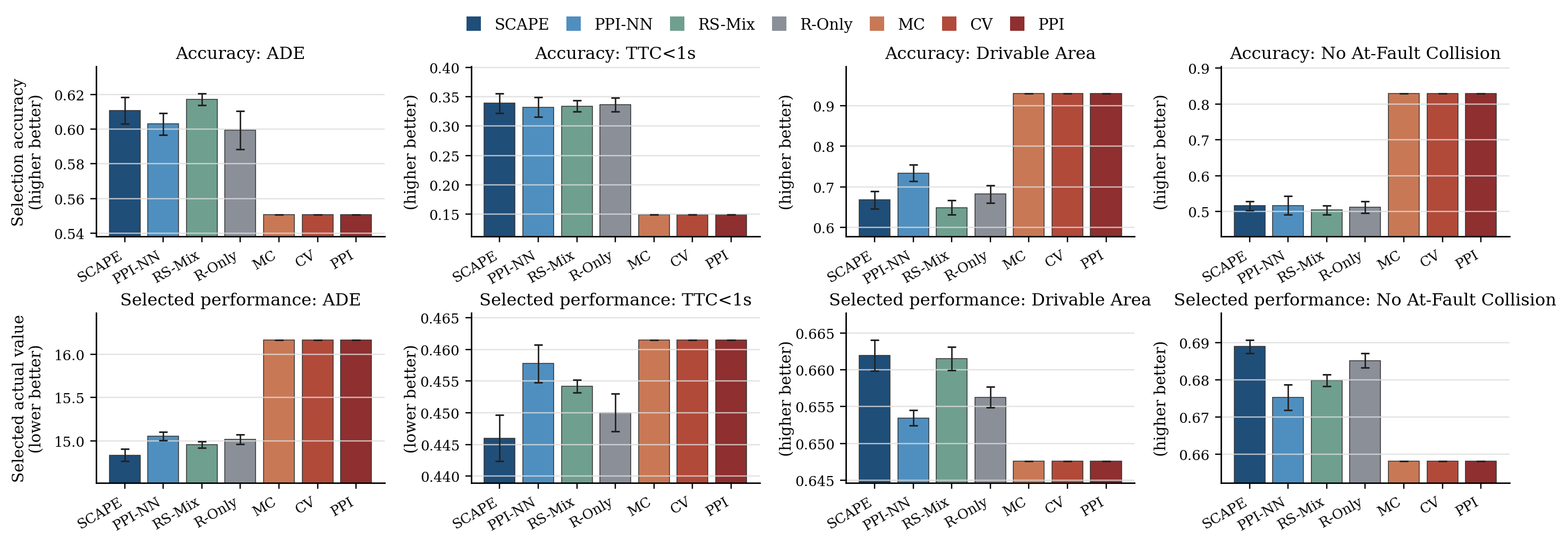}
        \caption{\(\rho_p=0.2\)}
        \label{fig:policy_moe_frac_02}
    \end{subfigure}

    \vspace{2pt}

    \begin{subfigure}[t]{0.48\linewidth}
        \centering
        \includegraphics[width=\linewidth]{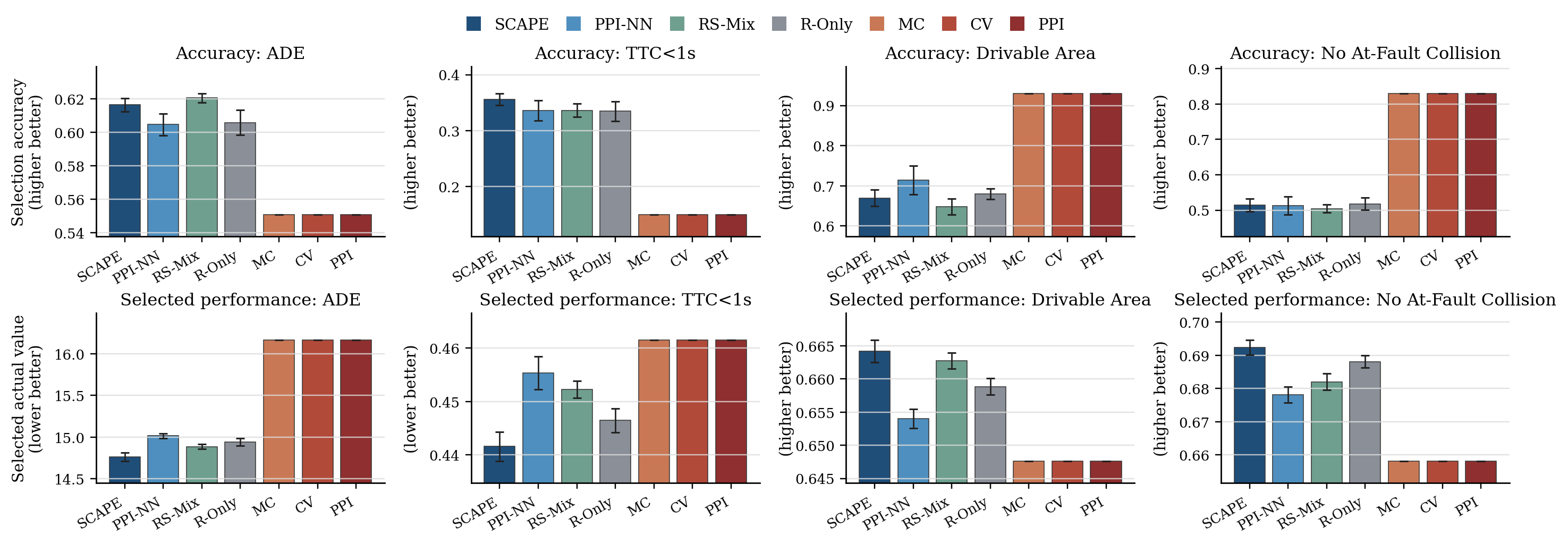}
        \caption{\(\rho_p=0.3\)}
        \label{fig:policy_moe_frac_03}
    \end{subfigure}
    \hfill
    \begin{subfigure}[t]{0.48\linewidth}
        \centering
        \includegraphics[width=\linewidth]{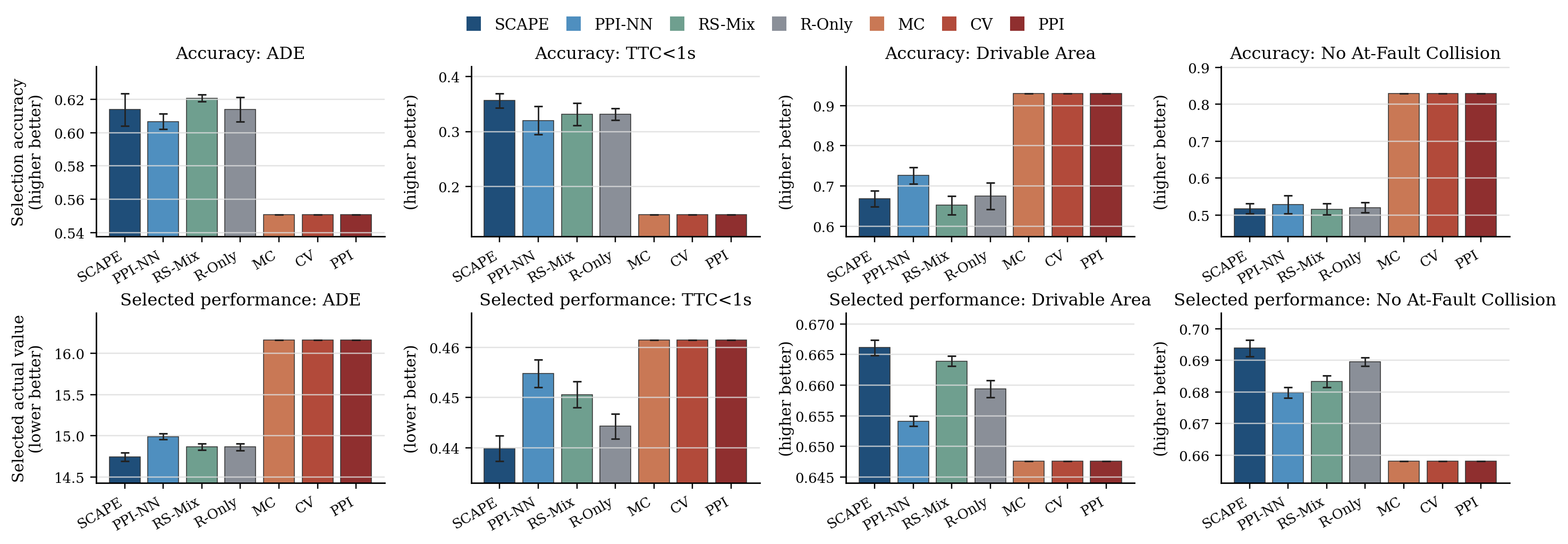}
        \caption{\(\rho_p=0.4\)}
        \label{fig:policy_moe_frac_04}
    \end{subfigure}

    \vspace{2pt}

    \begin{subfigure}[t]{0.48\linewidth}
        \centering
        \includegraphics[width=\linewidth]{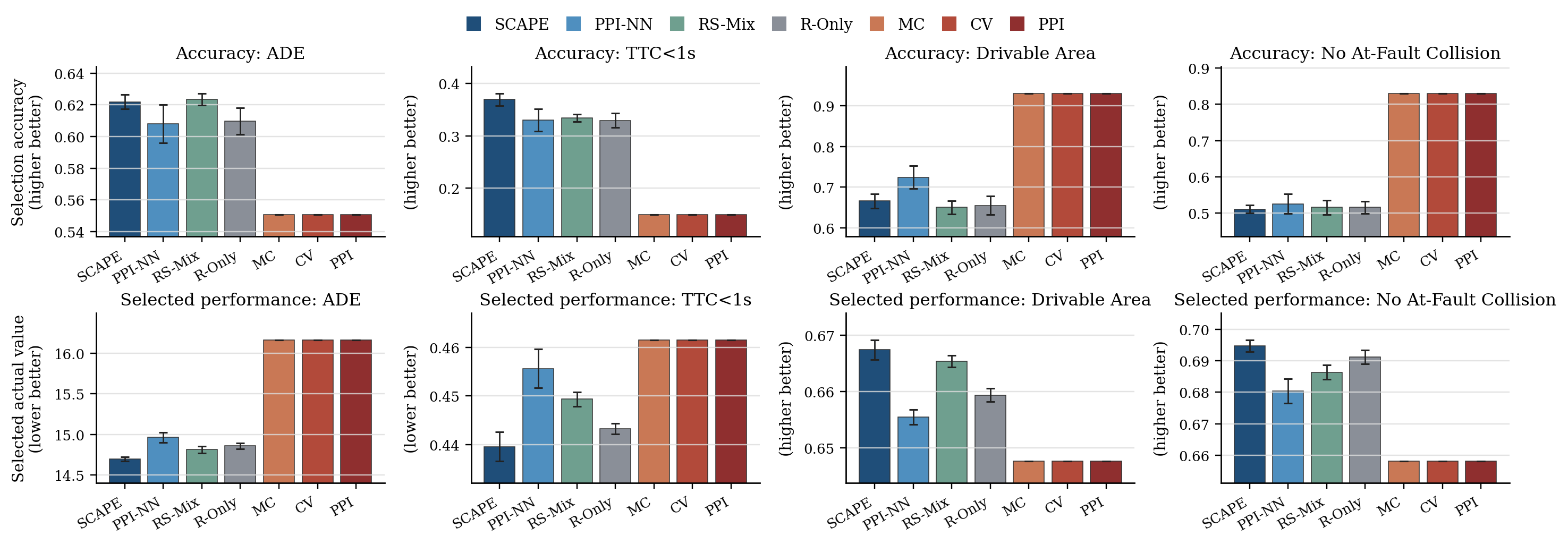}
        \caption{\(\rho_p=0.5\)}
        \label{fig:policy_moe_frac_05}
    \end{subfigure}
    \hfill
    \begin{subfigure}[t]{0.48\linewidth}
        \centering
        \includegraphics[width=\linewidth]{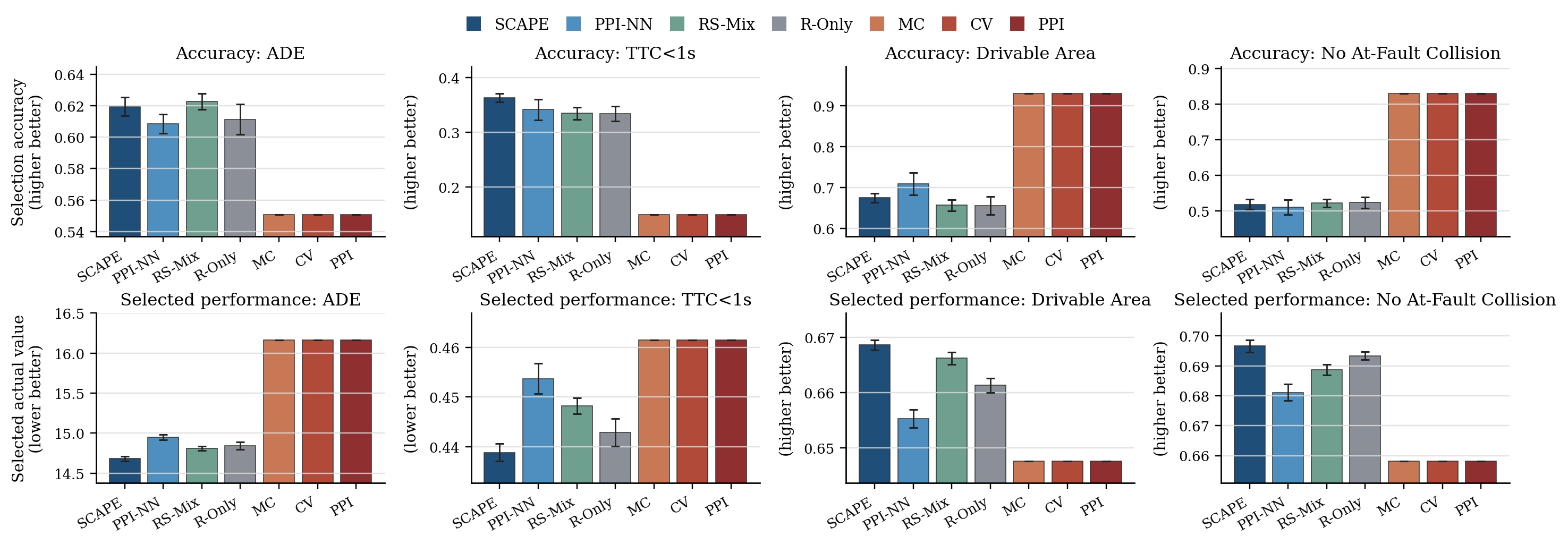}
        \caption{\(\rho_p=0.6\)}
        \label{fig:policy_moe_frac_06}
    \end{subfigure}

    \vspace{2pt}

    \begin{subfigure}[t]{0.48\linewidth}
        \centering
        \includegraphics[width=\linewidth]{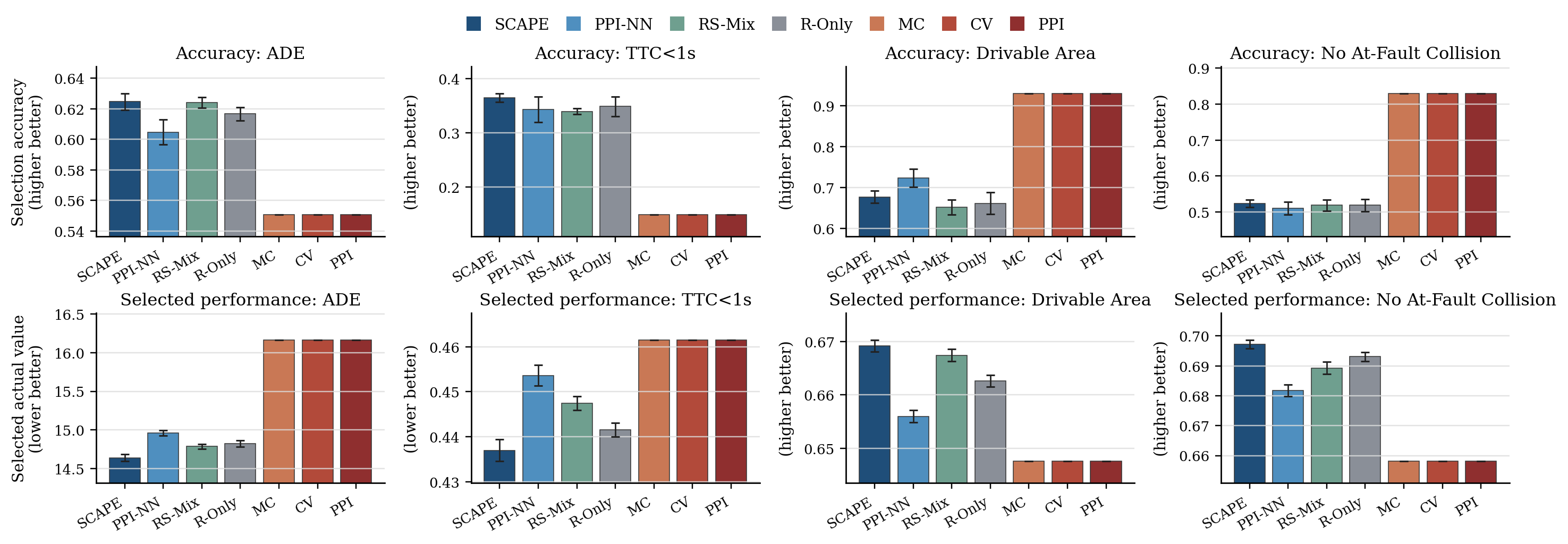}
        \caption{\(\rho_p=0.7\)}
        \label{fig:policy_moe_frac_07}
    \end{subfigure}
    \hfill
    \begin{subfigure}[t]{0.48\linewidth}
        \centering
        \includegraphics[width=\linewidth]{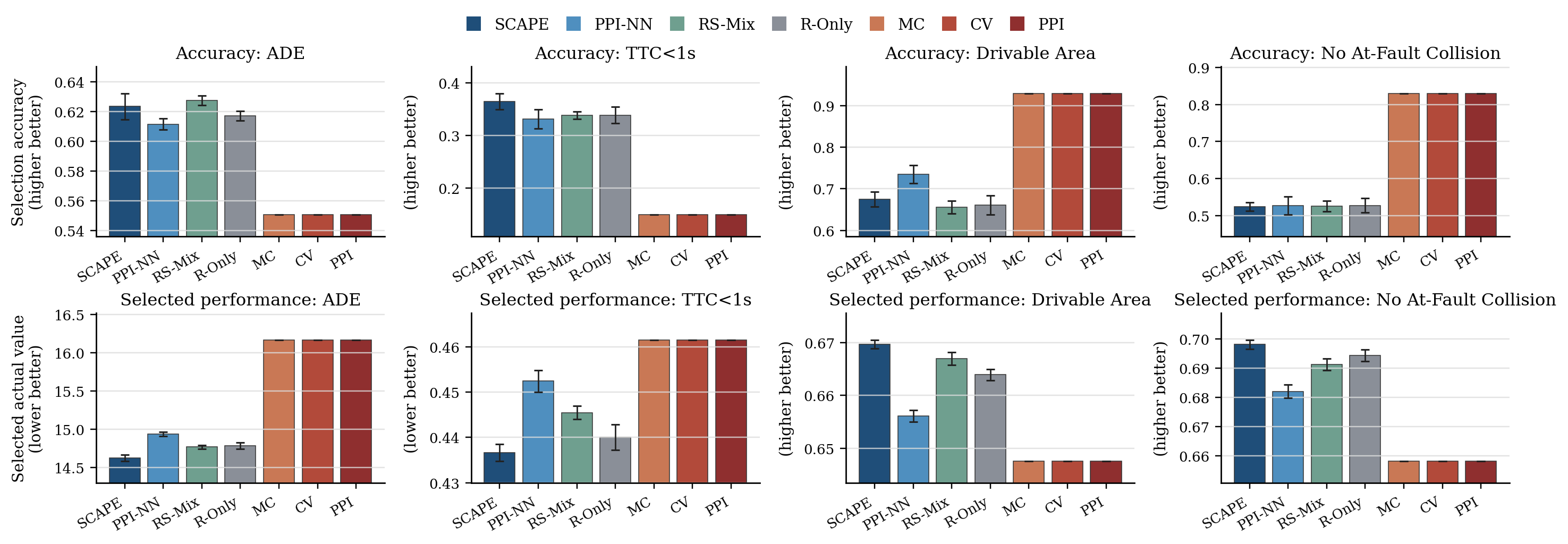}
        \caption{\(\rho_p=0.8\)}
        \label{fig:policy_moe_frac_08}
    \end{subfigure}

    \vspace{2pt}

    \begin{subfigure}[t]{0.48\linewidth}
        \centering
        \includegraphics[width=\linewidth]{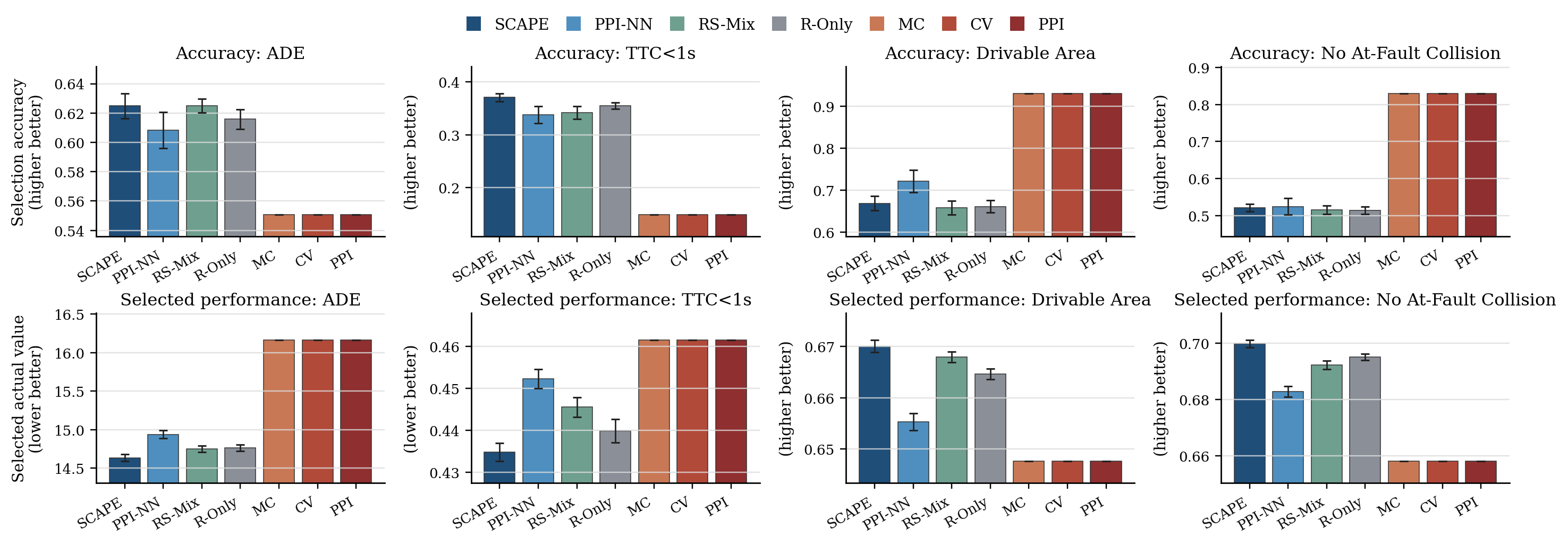}
        \caption{\(\rho_p=0.9\)}
        \label{fig:policy_moe_frac_09}
    \end{subfigure}

    \caption{\small \textbf{Additional evaluator-guided planner selection results on nuPlan.}
    Results are shown across paired-data fractions \(\rho_p\in\{0.1,\ldots,0.9\}\), where \(\rho_p\) denotes the fraction of the full paired training set \(\mathcal{D}_p\) used for training.
    Bars report mean \(\pm\) standard error over 10 paired-data seeds.}
    \label{fig:policy_moe_all_fracs}
    \vspace{-8pt}
\end{figure}

\end{document}